\documentclass[conference]{IEEEtran}
\IEEEoverridecommandlockouts
\usepackage{cite}
\usepackage{amsmath,amssymb,amsfonts}
\usepackage{graphicx}
\usepackage{textcomp}
\usepackage{xcolor}
\usepackage{listings}
\usepackage{algpseudocode}
\usepackage{algorithm}

\usepackage{babel,adjustbox,multirow}
\usepackage{booktabs}
\usepackage{cuted}
\def\BibTeX{{\rm B\kern-.05em{\sc i\kern-.025em b}\kern-.08em
    T\kern-.1667em\lower.7ex\hbox{E}\kern-.125emX}}
\begin{document}

\title{Long Range Switching Time Series Prediction \\ via State Space Model}

\author{\IEEEauthorblockN{1\textsuperscript{st} Jiaming Zhang\textsuperscript{1}}
\IEEEauthorblockA{\textit{Department of Computing} \\
\textit{Imperial College London}\\
London SW7 2AZ, United kingdom \\
\textit{School of Mathematics \& Statistics} \\
\textit{University of Glasgow} \\
Glasgow G12 8QQ, United Kingdom \\
jiaming.zhang@glasgow.ac.uk}
\and
\IEEEauthorblockN{2\textsuperscript{nd} Yang Ding\textsuperscript{1}}
\IEEEauthorblockA{\textit{Business School} \\
\textit{University of Edinburgh}\\
Edinburgh EH8 9JS, United Kingdom \\
\textit{Edinburgh Futures Institute} \\
\textit{University of Edinburgh} \\
Edinburgh EH3 9EF, United Kingdom\\
yang.ding@ed.ac.uk}
\and
\IEEEauthorblockN{3\textsuperscript{rd} Yunfeng Gao\textsuperscript{2}}
\IEEEauthorblockA{\textit{Business School} \\
\textit{Guilin University of Electronic Technology}\\
Guilin, China \\
yf\_yfg@163.com}
}

\maketitle

\begin{abstract}
In this study, we delve into the Structured State Space Model (S4), Change Point Detection methodologies, and the Switching Non-linear Dynamics System (SNLDS). Our central proposition is an enhanced inference technique and long-range dependency method for SNLDS. The cornerstone of our approach is the fusion of S4 and SNLDS, leveraging the strengths of both models to effectively address the intricacies of long-range dependencies in switching time series. Through rigorous testing, we demonstrate that our proposed methodology adeptly segments and reproduces long-range dependencies in both the 1-D Lorenz dataset and the 2-D bouncing ball dataset. Notably, our integrated approach outperforms the standalone SNLDS in these tasks.
\end{abstract}


\footnotetext[1]{~Jiaming Zhang and Yang Ding contributed to this article equally.}
\footnotetext[2]{~Corresponding author.}

\section{Introduction}
Generative modelling represents a rapidly advancing domain within the realm of artificial intelligence and machine learning. Its core focus lies in crafting models with the ability to generate data resembling real-world instances. The historical trajectory of generative modelling can be traced from its origins in rule-based systems and expert systems to its contemporary manifestation through modern deep learning methodologies.

Initial endeavours during the 1960s and 1970s predominantly revolved around Markov models and Hidden Markov Models (HMMs), which laid the cornerstone for sequence generation. The subsequent era, specifically the 1990s, saw the advent of Long Short-Term Memory (LSTM)\cite{b1}, a variant of Recurrent Neural Network meticulously designed to address the challenge of learning long-term dependencies in sequence prediction tasks.

However, the transformative breakthroughs of the 2010s catapulted generative modelling into the forefront of research. Variational Autoencoders (VAEs)\cite{b2} and Generative Adversarial Networks (GANs)\cite{b3} played pivotal roles in driving substantial advancements in diverse fields, including image synthesis and text generation.

Subsequently, the epochal introduction of the Transformer architecture\cite{b4},  further heightened the landscape. This groundbreaking neural network showcased its prowess in analysing extensive datasets at scale, thereby automating the creation of Large Language Models (LLMs). In 2020, the stage welcomed the emergence of Diffusion probabilistic models\cite{b5}, a latent variable framework drawing inspiration from non-equilibrium thermodynamics, further intensifying the competitive arena of generative modelling. 

Nonetheless, effectively modelling long sequences remains a formidable conundrum within the realm of machine learning. Crucial information required for task resolution could potentially be intricately intertwined across observations spanning thousands of temporal steps. Variant of Recurrent Neural Networks (RNNs) (\hspace{-0.5pt}\cite{b6},\cite{b7},\cite{b8}), Convolutional Neural Networks (CNNs) (\hspace{-0.5pt}\cite{b9},\cite{b10},\cite{b11}), and transformers (\hspace{-0.5pt}\cite{b4}) have been meticulously devised in an attempt to tackle this challenge. Specifically, numerous advanced transformer techniques have been developed (\hspace{-0.5pt}\cite{b12},\cite{b13},\cite{b13},\cite{b15}) to address the standard transformer's quadratic complexity relative to sequence length. Yet, these optimised transformers often struggle on extremely long-range sequence tasks\cite{b16}. 

Albert\cite{b17} introduced a novel method employing Structured State Space Sequence (S4) layers. An S4 layer facilitates a nonlinear sequence-to-sequence transformation through a collection of independent single-input single-output (SISO) linear state space models (SSMs), as described by Albert\cite{b18}. These SSMs are interconnected using nonlinear mixing layers. The HiPPO framework (\hspace{-0.5pt}\cite{b19}) is instrumental in these SSMs, with state matrices initialised in a specific manner. Since the SSMs are linear, each layer can be equivalently implemented as convolutions, enabling efficient parallel processing across sequence lengths. By stacking multiple S4 layers, one can design a deep sequence model. These models have notably outperformed earlier methods, especially in benchmarks like the Long Range Arena (LRA)\cite{b16} which aims to challenge long-range sequence models. Additionally, subsequent adaptations have demonstrated impressive results in tasks such as raw audio generation\cite{b20} and extended movie clip classifications\cite{b21}. 

A Switching Linear Dynamical System model (SLDS) captures a complex system's non-linear dynamic behaviour by transitioning between a suite of either linear dynamic models or non-linear dynamic models over time. Dong\cite{b22} proposed an novel methods Switching Non-Linear Dynamical System model (SNLDS) that can segment time series of high-dimensional signals into discrete temporal ``modes'' which will helps to determine the discrete hidden state is changed and impacts to the continues hidden state and also the observation.

S4 does not perform well in the switching time series, Although it can capture the whole history of the input signal, but it does not understand when to change the state as S4 is a linear transformation as it cannot hold the switching linear dynamics as it consist by multiple linear or non-linear dynamics. Also SNLDS does not have the long range dependency property for generative modelling.

Recognising these challenges, our approach synergies the robustness of S4 with the flexibility of switching linear dynamics, culminating in the integrated S4 SNLDS model.

The following sections of the study are organised as follows. Section \ref{sec:background} consists the background of S4 models, change point detection methods and SNLDS models. Section \ref{sec:S4_split} provides 
a novel model S4 split that combines S4 and Change Finder algorithm together with experiments to prove the result.
Section \ref{S4_SNLDS} introduces another model S4 SNLDS that overcome the disadvantage of S4 split and combine S4 and SNLDS together that can generative long range switching data without lost of accuracy. Section \ref{Conclusion} concludes the work done in this study and future work.

\section{Background}
\label{sec:background}

\subsection{State Space Models}

State Space Models (SSMs) are frameworks used to study deterministic and stochastic dynamical systems. The Kalman filter\cite{b23} is a popular SSM used for estimating and predicting system states based on noisy measurements. Hidden Markov Models (HMMs) are a type of SSM that model time series or sequence data assuming a Markov process with hidden or unobserved states. Both the Kalman filter and HMMs provide a way to predict sequential models. SSMs' hidden state and input transition matrices can be learned directly by gradient descent to model sequences of observations\cite{b18}. In a seminal work, \cite{b17} demonstrated that SSMs can be made into the most potent sequence modelling framework by incorporating algorithmic methods for memorising and computing input sequences.

A continuous time latent space model maps a $1$-dimensional input sequence $u(t)$ to a $N$-dimensional latent state $x(t)$, after which $x(t)$ is mapped to a $1$-dimensional output sequence $y(t)$.
\begin{equation}
  \label{eq:ssm}
  \begin{aligned}
    \dot{x}(t) &= \textbf{A}x(t) + \textbf{B}u(t), \\
    y(t) &= \textbf{C}x(t) + \textbf{D}u(t).
  \end{aligned}
\end{equation}
Where $\textbf{A} \in \mathbb{R}^{N \times N}$ is a state matrix, $\textbf{B} \in \mathbb{R}^{N \times 1}$ is the input matrix, $\textbf{C} \in \mathbb{R}^{1 \times N}$ is the output matrix and $\mathbf{D} \in \mathbb{R}$ is the feed through matrix.

\subsection{The HiPPO framework}

Albert\cite{b19} introduced the High-order Polynomial Projection Operators (HiPPO) to address issues such as the vanishing gradient problem and limited memory horizon problem in RNNs. Additionally, the Legendre Memory Units (LMU)\cite{b24} has been proposed as a novel memory cell that can dynamically maintain information across long-time windows using relatively few resources, which speeds up a specific linear recurrence using convolutions. It uses mathematical principles to orthogonalise its continuous-time history by solving $d$ coupled ordinary differential equations (ODEs). These ODEs have a phase space that is linearly mapped onto sliding windows of time using Legendre polynomials of degree $d-1$. But LMU requires priors on the sequence length or timescale and is ineffective outside this range. Hence, HiPPO framework was inspired by LMU and proposed to improve such problems and generalise the LMU into a theoretical framework for continuous-time memorisation. 

The HiPPO framework is a method for projecting functions onto a space of orthogonal polynomials with respect to a given measure. By using this framework, it is possible to analyse a range of different measures and incrementally update the optimal polynomial approximation as the input function is revealed over time. The HiPPO framework can be implemented as Eq. \ref{eq:ssm}, allowing for fast updates and efficient analysis of the input function. The matrix $\mathbf{A}$ is then defined as HiPPO matrix. 

HiPPO defines a projection operator $\emph{proj}_t$ and a coefficient extraction operator $\emph{coef}_t$ that $\emph{proj}_t$ takes a function $f: \mathbb{R} \rightarrow \mathbb{R}$ map it to a polynomial $g \in G$, an N-dimensional subspace of polynomials, that minimises the approximation error, and $\emph{coef}_t$ maps the polynomial $g$ to the coefficients $c(t) \in \mathbb{R}^N$ that the basis of the orthogonal polynomials defined with respect to the measure $\mu$. The HiPPO opeartor is then defined as \emph{coef} $\circ$ \emph{proj}, which is an operator that map a function to the optimal projection coefficent, i.e., $ HiPPO(f(t))=\emph{coef}(\emph{proj}(f(t)))$.

The HiPPO framework yields a new memory update mechanism when we initialise HiPPO matrix $\mathbf{A}$ into a different form, i.e., HiPPO-LegS matrix \ref{legs} scales through time to remember all history, avoiding priors on the timescale, which avoids the memory issues in RNN. 
\begin{equation}
 \mathbf{A}_{n k}=- \begin{cases}(2 n+1)^{\frac{1}{2}}(2 k+1)^{\frac{1}{2}} & n>k \\
n+1 & n=k \\
0 & n<k\end{cases} 
\label{legs}
\end{equation}

\subsection{Linear State Space Layer}

Linear State Space Layer (LSSL)\cite{b18} leveraged the HiPPO framework and generalises convolutions to continuous-time, explains common RNN heuristics and shares features of Nerual Differential Equations (NDEs) such as time-scale adaptation. LSSL is defined by Eq. \ref{eq:ssm}.

LSSLs unify the strengths of Continuous-time, RNNs and CNNs models:
\begin{itemize}
    \item Continuous-time: The LSSL model is formulated as a differential equation, granting it the flexibility to handle even irregularly-spaced data. This adaptability is achieved by discretising the matrix $\mathbf{A}$ in accordance with a varying timescale \( \Delta t \).
    \item Recurrent: When a specific discrete step-size \( \Delta t \) is chosen, the model can be adapted into a recurrent format. This allows for efficient inference by sequentially computing each layer, made possible through the unrolling of the linear recurrence.
    \item Convolutional: Linear time-invariant systems, as described by equation \eqref{eq:ssm}, can be explicitly mapped to a continuous convolution operation.
\end{itemize}

To operate on discrete-time sequences sampled with a \textbf{step size} $\Delta$, SSMs can be computed with the recurrence via Approximations of differential equations and Generalised Bilinear transform (GBT)
\begin{equation}
\label{eq:ssm-discrete}
\begin{aligned}
  x_{k} &= \mathbf{\overline{A}} x_{k-1} + \mathbf{\overline{B}} u_k &
  \mathbf{\overline{A}}  &= (\mathbf{I} - \Delta/2 \cdot \mathbf{A})^{-1}(\mathbf{I} + \Delta/2 \cdot \mathbf{A}), &
  \\
  y_k &= \mathbf{\overline{C}} x_k + \mathbf{{D}} u_k &
  \mathbf{\overline{B}} &= (\mathbf{I} - \Delta/2 \cdot \mathbf{A})^{-1} \Delta \mathbf{B}, \mathbf{\overline{C}} = \textbf{C}.
\end{aligned}
\end{equation}
Where $\mathbf{\overline{A}}, \mathbf{\overline{B}}$ are the discretized state matrix. Then unrolling Eq. \ref{eq:ssm-discrete}, we can derive 
\begin{equation}
  \label{eq:convolution}
  y_k = \mathbf{\overline{C}}\mathbf{\overline{A}}^k \mathbf{\overline{B} }u_0 + \mathbf{\overline{C}}\mathbf{\overline{A}}^{k-1} \mathbf{\overline{B}} u_1 + \dots + \mathbf{\overline{C}} \mathbf{\overline{A} \overline{B}} u_{k-1} + \mathbf{\overline{B}} u_k
  + \mathbf{D} u_k
  .
\end{equation}

This can be written as a convolutional representation $y = \overline{\mathbf{K}} * u$, where the convolution kernel is
\begin{equation}%
  \label{eq:ssm-conv}
  \mathbf{\overline{K}} \in \mathbb{R}^L = (\mathbf{\overline{C}}\mathbf{\overline{B}}, \mathbf{\overline{C}}\mathbf{\overline{A}}\mathbf{\overline{B}}, \dots, \mathbf{\overline{C}}\mathbf{\overline{A}}^{L-1}\mathbf{\overline{B}}).
\end{equation}

While the LSSL model offers theoretical advantages, it remains impractical for real-world applications due to the computational and memory burdens imposed by its state representation. To address these challenges, a subsequent model has been proposed, the S4. S4 offers innovative parameterisation and algorithms specifically designed to make state space modelling more computationally efficient and memory-friendly.

\subsection{Structured state space sequence model}
Structured state space sequence model (S4) uses a structural result that puts HiPPO matrix $\mathbf{A}$ into a canonical form by conjugation, which is ideally more structured and allows faster computation\cite{b17}. Hence S4 reparameterises the structured state HiPPO matrices A (\ref{nplr}) by decomposing them as the sum of a low-rank and normal term which is called Normal Plus Low-Rank (NPLR). The HiPPO matrices A can have NPLR representation as 
\begin{equation}
    \label{eq:nplr}
    \mathbf{A} = \mathbf{V} \mathbf{\Lambda} \mathbf{V}^* - \mathbf{P} \mathbf{Q}^\top = \mathbf{V} \left( \mathbf{\Lambda} - \left(\mathbf{V}^* \mathbf{P}\right) (\mathbf{V}^*\mathbf{Q})^* \right) \mathbf{V}^*
\end{equation}
for unitary \( \mathbf{V} \in \mathbb{C}^{N \times N} \), diagonal \( \mathbf{\Lambda} \), and low-rank factorisation \( \mathbf{P}, \mathbf{Q} \in \mathbb{R}^{N \times r} \). In here the NPLR matrix $\mathbf{A}$ can be conjugated into diagonal plus low-rank (DPLR) form. 
Then we can use DPLR to reconstruct $\mathbf{A}$ as $(\mathbf{\Lambda} - \mathbf{P}\mathbf{Q}^*)$. However, this may suffer from numerical instabilities when the $\mathbf{A}$ matrix has positive real value eigenvalues\cite{b20}, hence we use $\mathbf{A} =  (\mathbf{\Lambda} - \mathbf{P}\mathbf{P}^*)$.
Then we apply Woodbury identity and Cauchy kernel to reduce the computational cost and increase the stability of computing the convolutional kernel $\hat{\mathbf{K}}$.
\begin{equation}
    \mathbf{A}_{n k}^{(N)}=- \begin{cases}\left(n+\frac{1}{2}\right)^{1 / 2}\left(k+\frac{1}{2}\right)^{1 / 2} & n>k \\
    \frac{1}{2} & n=k \\
    \left(n+\frac{1}{2}\right)^{1 / 2}\left(k+\frac{1}{2}\right)^{1 / 2} & n<k\end{cases} \\
\label{nplr}
\end{equation}
\begin{equation}
    \mathbf{B}_{n} = (2n+1)^{\frac{1}{2}}
    \label{B}
\end{equation}

The general architecture of S4, a deep neural network (DNN), is similar to that of previous models. Specifically, S4 operates as a sequence-to-sequence mapper with dimensions (batch size, sequence length, hidden dimension), aligning it with existing sequence models like RNNs, CNNs, and Transformers.

However, S4 distinguishes itself as a specialized hybrid of both CNNs and RNNs, inheriting the strengths and mitigating the weaknesses of each. From the RNN perspective, Equation \(\ref{eq:ssm-discrete}\) outlines a streamlined structure for S4 that sidesteps the optimization challenges commonly associated with traditional RNNs. On the CNN front, Equation \(\ref{eq:ssm-conv}\) describes an unbounded convolutional kernel for S4, effectively bypassing the context size limitations constraining standard CNNs.

Thus, S4 benefits from the computational efficiency of existing sequence models in training and inference and addresses their individual limitations, making it a robust and versatile model for sequence analysis.

\subsection{S4D}

The implementation of S4 as a deep learning model involves a sophisticated algorithm replete with intricate linear algebraic techniques, making it challenging both to understand and to implement. 
These complexities stem largely from the model's state matrix, which is parameterised as a Diagonal Plus Low-Rank (DPLR) matrix capable of capturing HiPPO matrices.
 Given these complexities, a natural problem arises regarding whether there are ways to simplify this parameterisation and its corresponding algorithm. Albert proposed a new method S4D\cite{b25} that removing the low-rank term would result in a diagonal state space model (DSS), offering a significantly more straightforward implementation and conceptual understanding. The core result is that instead of using DPLR matrices, S4D uses Vandermonde matrix multiplication to compute the matrix $\mathbf{A}$.  
 
 When $\mathbf{A}$ is diagonal, the computation is nearly trivial. By Eq. \ref{eq:ssm-conv}, we have
 \begin{equation}
 \begin{aligned}
     \overline{\mathbf{K}}_{\ell}=\sum_{n=0}^{N-1} \mathbf{C}_n \overline{\mathbf{A}}_n^{\ell} \overline{\mathbf{B}}_n \Longrightarrow &\overline{\mathbf{K}}=\left(\overline{\mathbf{B}}^{\top} \circ \mathbf{C}\right) \cdot \mathcal{V}_L(\overline{\mathbf{A}}) \\
     & \text { where } \mathcal{V}_L(\overline{\mathbf{A}})_{n, \ell}=\overline{\mathbf{A}}_n^{\ell}
 \end{aligned}
 \end{equation}
 where $\circ$ is Hadamard product, · is matrix multiplication, and $\mathcal{V}$ is known as a Vandermonde matrix.  

This outcome enables S4D to feature a straightforward, interpretable convolution kernel $\mathbf{K}$ that can be implemented with just two lines of code, i.e.,
\begin{lstlisting}[language=Python]
    def S4d_kerenal(A,B,C,dt,L):
        Vand = exp(arange(L)[:,None]*dt*A)
    return sum(Vand*B*C(exp(dt*A)-1),-1)
\end{lstlisting}

Moreover, S4D further simplified the initialisation of HiPPO matrix $\mathbf{A}$ into S4D-Inv
\begin{equation}
    \mathbf{A_n} = -\frac{1}{2} + i\frac{N}{\pi}\left( \frac{N}{2n+1} -1 \right)
    \label{s4d-inv}
\end{equation}
and the corresponding matrix $\mathbf{B}$ set to $\mathbf{1}$. S4D-Inv is the approximation of the S4 $\mathbf{A}$ matrix (\ref{nplr}). This is because DSS is a variant of S4, which is a noisy approximation to S4 $\mathbf{A}$ matrix (\ref{nplr})\cite{b25}.

\subsection{FlashConv}

To improve the efficiency of SSMs on GPUs, Daniel proposed a novel method, FlashConv\cite{b26}. FlashConv combined the Fast Fourier Transform (FFT), pointwise multiply, and inverse FFT to minimise memory reads/writes. It also employs a block FFT algorithm to leverage specialised matrix multiply units, but it only works for shorter sequence lengths (depending on GPU, here we are using A6000, which can hold up to 6k). For sequences longer than 6k, the computations are too large to fit in GPU, so Daniel proposes a novel State-Passing algorithm that splits the sequence into manageable chunks, allowing for chunk-by-chunk computation of the FFT convolution.

For sequences length shorter than 6k, Daniel apples kernel fusion and block FFT techniques to speed up the FFT-based convolution, which called  Fused Block FFT convolution. Kernel fusion solves the memory issue in the GPU due to the reading and writing of intermediate results. At the same time,  block FFT allows the FFT-based convolution to utilise specialised matrix multiplication units. Kernel fusion using FlashAttention\cite{b27} techniques to fuse the entire FFT convolution into a single kernel; this solves the memory issues in GPU. Block FFT means that FFT can be written as a series of block-diagonal matrix multiplications interleaved with permutation.  

The State-Passing algorithm can accelerate computing FFT convolution for long sequences as the fused kernel cannot fit into normal GPU memory.

The inherent recurrence in SSM enables  us to divide an FFT convolution of a sequence with length $N$ into manageable chunks, each chunks is size $N'$ ($N'$ is the largest FFT size that can be accommodated within the GPU SRAM), this approach assumes that $N$ is a multiple of $N'$. Then apply Fused Block FFT convolution to compute each chunks individually and employ a recurrence relation to connect them together coherently. Algorithm \ref{alg:fft} shows the State-Passing algorithm for 1D input.

\begin{algorithm}
\caption{State-Passing algorithm From \cite{b26}}\label{alg:fft}
\begin{algorithmic}[1]
\State{Input: $ u \in \mathbb{R}^N, \mathrm{SSM} \text { parameterised by matrices } \mathbf{A} \in \mathbb{R}^{m \times m}, \mathbf{B} \in \mathbb{R}^{m \times 1}, \mathbf{C} \in \mathbb{R}^{1 \times m}, \mathbf{D} \in \mathbb{R}^{1 \times 1}, \text { chunk size } N^{\prime} \text { where } N \text { is a multiple of } N^{\prime} . $}
\State{Precompute $\mathbf{A}^{N^{\prime}} , \mathbf{M}_{u x}=[\mathbf{A}^{N^{\prime}-1} \mathbf{B}, \ldots, \mathbf{B}] \in \mathbb{R}^{m \times N^{\prime}}, \mathbf{M}_{x y}=[\mathbf{C}, \ldots, \mathbf{C A}^{N^{\prime}-1}] \in \mathbb{R}^{N^{\prime} \times m} .$}
\State{$\text {Split the inputs } u_{1: N} \text { into } C=N / N^{\prime} \operatorname{chunks} u_{1: N^{\prime}}^{(c)} \text { for } c=1, \ldots, C . $}
\State{$\text {Let the initial state be } x_{N^{\prime}}^{(0)}=0 \in \mathbb{R}^m .$}
\For{$1 \leq c \leq C$}
    \State{$\text { Compute } y^{(c)}=\mathbf{M}_{x y} x_{N^{\prime}}^{(c-1)}+\operatorname{Fused Block FFT convolution}\left(f, u_j\right)+\mathbf{D} u^{(c)} \in \mathbb{R}^{N^{\prime}} .$}
    \State{$\text {Update state: } x_{N^{\prime}}^{(c)}=\mathbf{A}^{N^{\prime}} x_{N^{\prime}}^{(c-1)}+\mathbf{M}_{u x} u^{(c)} .$}
\EndFor
\State{Return: $y=\left[y^{(1)} \ldots y^{(C)}\right]$}
\end{algorithmic}
\end{algorithm}

\subsection{Change-point Detection Method}

In statistical analysis and data mining, change-point detection has been used for various purposes, such as step detection, edge detection, and anomaly detection. Sequential Discounting Auto Regression learning (SDAR) algorithm\cite{b28} is an efficient and simple algorithm to identify the outliers from the data. The change finder algorithm employs the SDAR algorithm. It consists of two learning phases, outlier detection and change point detection.

Let a mean of an initial value is zero time series as ${z_t: t=1,2, \dots}$, where $z(t) \in \mathbb{R}^d$. An AR model of the $k$ th order is given by
$$
    z_t=\sum_{i=1}^k A_i z_{t-i}+\varepsilon,
$$
where each $A_i \in \mathbb{R}^{d \times d}$, $\varepsilon \sim \mathcal{N}(0, \Sigma)$. Let ${x_t: t=1,2, \dots}$ denote a time series, where
$$
x_t=z_t+\mu .
$$
Then, let $x_{t-k}^{t-1}=x_{t-k} \cdots x_{t-1}$, the conditional probability density function of $x_t$ given $x_{t-k}^{t-1}$ is denoted as 
$$
p\left(x_t | x_{t-k}^{t-1}: \theta\right)=  \frac{1}{(2 \pi)^{k / 2}|\Sigma|^{1 / 2}} \exp \left(-\frac{1}{2}\left(x_t-w_t\right)^T \Sigma^{-1}\left(x_t-w_t\right)\right),
$$
where
$$
w_t=\sum_{i=1}^k A_i\left(x_{t-i}-\mu\right)+\mu .
$$
We set $\theta=\left(A_1, \cdots, A_k, \mu, \Sigma\right)$. Let $\theta_t$ denote an estimate of $\theta$ given $x^t$, then  $p_t=p\left(\cdot | \cdot, \theta_t\right)$. For an estimation of $\theta$,  employ maximum likelihood method to computes the value of $\theta$ maximising the following quantity:
$$
\sum_{i=1}^t(1-r)^{t-i} \log p\left(x_i | x^{i-1}, \theta\right)
$$
in an online manner. This is the SDAR algorithm.

At the first stage, outlier detection, the algorithm constructs constructs a sequence of probability density functions denoted as ${p_t(x): t = 1,2,\dots}$ that characterises the underlying dynamics of the data generation process. We assume that each $p_t$ represents a density of stochastic process. For a stochastic process $p$, the conditional probability density function of \( x_{t+1} \) given the sequence \( x^t = x_1, x_2, \ldots, x_t \) is denoted by \( p(x_{t+1} | x^t) \). Employ SDAR algorithm to estimate the stochastic process $p$. For each input $x_t(t \geq k+1)$,  the algorithm calculate the score of $x_t$ using the following formula:
\begin{equation}
{Score}\left(x_t\right)=-\log p_{t-1}\left(x_t | x^{t-1}\right),
\label{score}
\end{equation}
where  Eq. \ref{score} shows how the prediction loss for $x_t$ is related relative to a probability density function $p_{t-1}\left(\cdot | x^{t-1}\right)$.

Then in the second stage, the change point detection. Let $T$ be a positive constant and $\left\{x_t\right\}$ be a data sequence, define $y_t$ as the T-averaged score over $\left\{{Score}\left(x_i\right): i=t-T+1, \dots, t\right\}$ as:
\begin{equation}
    y_t=\frac{1}{T} \sum_{i=t-T+1}^t {Score}\left(x_i\right),
    \label{eq_y}
\end{equation}
where \( \text{Score}(x_i) \) is calculated according to Eq. \(\ref{score}\). From this, we generate a time series \( y_t \) for \( t=1,2,\ldots \). For \( t \leq T \), we compute \( y_t \) using dummy values for \( \text{Score}(x_i) \) where \( i \leq T \). This step is essential for mitigating the impact of isolated outliers in the sequence \( \{x_t\} \).

If a change point occurs at time \( t = t_1 \), we expect to observe elevated outlier scores for a period following \( t_1 \). However, if \( t_1 \) is merely an isolated outlier rather than a change point, then the elevated score will be limited to \( t_1 \) itself. This isolated high score can be smoothed out via the moving average process defined in Eq. \(\ref{eq_y}\).

Subsequently, the SDAR algorithm is employed to learn from the time series \( \{y_t\} \), resulting in a series of AR models denoted as \( \{q_t\} \).

The averaged score over a window of \( T \) time steps at time \( t \) is defined in Eq. \(\ref{score}\) as:
\begin{equation}
    {Score}(t)=\frac{1}{T} \sum_{i=t-T+1}^t\left(-\log q_{i-1}\left(y_i | y_{i-k^{\prime}}^{i-1}\right)\right) .
    \label{eq-4}
\end{equation}
This equation averages the log-likelihoods over a window, effectively serving as a smoothed measure of anomaly or fit within the observed time series.

This approach fundamentally reframes change point detection as an outlier detection problem using time series outlier scores, leveraging two moving average processes described in Eq. \ref{eq_y} and Eq. \ref{eq-4}. This unifies outlier detection and change point detection within a cohesive framework, emphasising their intrinsic linkage. A higher $S c o r e(t)$ suggests a greater likelihood that time point $t$ is a change point.

The two moving average processes primarily serve to mitigate the impact of individual outliers. When $T$ is small, both outliers and change points are detected almost instantaneously upon their occurrence, though distinguishing between the two can pose challenges. Conversely, with a larger $T$, there's an inherent delay in detecting change points, but this ensures that only the outliers are filtered out, leading to more precise detection of significant change points.

\subsection{Switching Non-linear Dynamics System}
\label{sec:snlds}
The state space model is given by:
\begin{equation}
\label{generative}
\begin{aligned}
    & p_\theta(\mathbf{x}, \mathbf{z}, \mathbf{s}) = \\
    & p\left(\mathbf{x}_1 | \mathbf{z}_1\right) p\left(\mathbf{z}_1 | s_1\right) {\left[\prod_{t=2}^T p\left(\mathbf{x}_t | \mathbf{z}_t\right) p\left(\mathbf{z}_t | \mathbf{z}_{t-1}, s_t\right) p\left(s_t | s_{t-1}, \mathbf{x}_{t-1}\right)\right] }
\end{aligned}
\end{equation}
where, \(s_t \in \{1, \ldots, K\}\) represents the discrete hidden state, \(\mathbf{z}_t \in \mathbb{R}^L\) the continuous hidden state, and \(\mathbf{x}_t \in \mathbb{R}^D\) the observed output. For notation simplicity, we ignore the input $\mathbf{u}_t$.

Notably, the discrete state \(s_t\) does not only depend on the previous discrete state \(s_{t-1}\), as one would see in a Hidden Markov Model (HMM) but also by the previous observation \(\mathbf{x}_{t-1}\). This feature allows for a more interactive or ``closed-loop'' state evolution, where the state can respond by signals from the environmental.

\begin{figure}[b]
    \centering
    \includegraphics[width=1\linewidth]{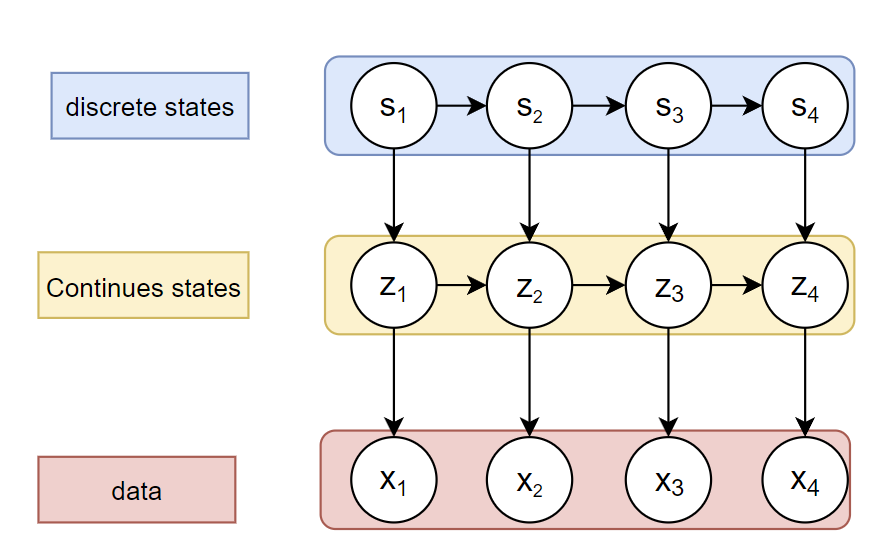}
    \caption{Illustration of the generative model.}
    \label{fig:snds-label}
\end{figure}

While the model is described under the assumption of first-order Markov dependencies for simplicity, extending it to accommodate higher-order dependencies is straightforward. For example, the continuous hidden state \(\mathbf{z}_t\) could be conditioned on the previous observation \(\mathbf{x}_{t-1}\), and the discrete hidden state \(s_t\) could depend on the previous continuous state \(\mathbf{z}_{t-1}\). Incorporating such additional dependencies is feasible and readily integrated into the inference framework, as indicated by the dashed lines in Fig. \ref{fig:snds-label}.

The conditional probability distributions are defined:
\begin{equation}
\label{snldsassump}
\begin{aligned}
p\left(\mathbf{x}_t | \mathbf{z}_t\right) &= \mathcal{N}\left(\mathbf{x}_t | f_x\left(\mathbf{z}_t\right), \mathbf{R}\right), \\
p\left(\mathbf{z}_t | \mathbf{z}_{t-1}, s_t=k\right) &= \mathcal{N}\left(\mathbf{z}_t | f_z\left(\mathbf{z}_{t-1}, k\right), \mathbf{Q}\right), \\
p\left(s_t | s_{t-1}=j, \mathbf{x}_{t-1}\right) &= \operatorname{Cat}\left(s_t | \mathcal{S}\left(f_s\left(\mathbf{x}_{t-1}, j\right)\right)\right),
\end{aligned}
\end{equation}
where \( f_{x}, f_{z}, f_{s} \) are nonlinear functions (e.g., Multi-Layer Perceptions (MLPs) or RNNs), \( \mathcal{N}(\cdot, \cdot) \) denotes a multivariate Gaussian distribution, \( \operatorname{Cat}(\cdot) \) is a categorical distribution, and \( \mathcal{S}(\cdot) \) is a softmax function. Additionally, \( \mathbf{R} \in \mathbb{R}^{D \times D} \) and \( \mathbf{Q} \in \mathbb{R}^{H \times H} \) are learned covariance matrices for the Gaussian emission and transition noise, respectively.

Special cases of this model can be described as follows:
\begin{itemize}
    \item Recurrent Switching Linear Dynamical System (SLDS): If both \( f_x \) and \( f_z \) are linear functions and the dependence of \( s_t \) on \( s_{t-1} \) is a first-order Markov process that also depends on \( z_{t-1} \), then the model is referred to as a recurrent SLDS \cite{b29}.
    \item Switching Linear Dynamical System (SLDS):  If \( f_x \) and \( f_z \) are linear, but \( s_t \) does not depend on \( z_{t-1} \), the model simplifies to a regular SLDS.
    \item Switching Nonlinear Dynamical System (SNLDS): If both \( f_x \) and \( f_z \) are nonlinear, then the model is characterised as a SNLDS.
\end{itemize}

These variations allow for a broad range of complexities and behaviours, which can be chosen to match the particularities of the application or the observed data.

\subsection{Variational inference and learning}

Variational inference aims to approximate the conditional density of latent variables based on observed ones. At its core, it addresses this through optimisation. By employing a set of densities over the latent variables, defined by adaptable ``variational parameters'', the optimisation process identifies the specific parameter configuration that minimises the KL divergence with the target conditional. This optimised variational density then acts as a stand-in for the precise conditional density.

Let \( \mathbf{x} = x_{1:n} \) be a set of observed variables, and \( \mathbf{z} = z_{1:m} \) be a set of latent (or hidden) variables. The joint density of these variables is \( p(\mathbf{z}, \mathbf{x}) \).

The conditional probability of the latent variables given the observed variables is defined as:
\begin{equation}
    p(\mathbf{z} | \mathbf{x}) = \frac{p(\mathbf{z}, \mathbf{x})}{p(\mathbf{x})}
\end{equation}
where the denominator, \( p(\mathbf{x}) \), is the evidence. The evidence is computed by marginalising the latent variables \( \mathbf{z} \) out of the joint distribution \( p(\mathbf{z}, \mathbf{x}) \):
\begin{equation}
    p( \mathbf{x})=\int p(\mathbf{z}  ,\mathbf{x}) dz.
    \label{eq:marginal}
\end{equation}

However, it is hard to compute $p(\mathbf{x})$ as the complexity is not manageable. Therefore we specify a family $\mathcal{Q}$ of densities over the latent variables. Each $q(\mathbf{z}) \in \mathcal{Q}$ is a candidate approximation to the exact conditional.

The marginal $p(\mathbf{x})$ above can be computed as:
\begin{equation}
p(\mathbf{x})=\iint_{\ldots} \int p\left(\mathbf{x}, z_1, z_2, z_3, \ldots\right) d z_1 d z_2 d z_3 \ldots
\end{equation}
In variational inference, we avoid computing the marginal $p(\mathbf{x})$ by  employing a tractable distribution \( q(\mathbf{z}) \) to approximate the exact conditional distribution \( p(\mathbf{z} | \mathbf{x}) \):
\begin{equation}
p\left(z_1, z_2, \ldots | \mathbf{x}\right) \approx q\left(z_1, z_2, \ldots | \theta\right)
\end{equation}
where \( \theta \) represents the parameters of the distribution \( q \), learned from the data. We choose \( q \) from a family of distributions that are computationally easier to work with, ensuring that key statistical measures like expectation and variance can be calculated directly.

In variational inference, the overarching aim is to identify an approximation \( q^*(\mathbf{z}) \) that is closest to the true posterior distribution \( p(\mathbf{z} | \mathbf{x}) \) in terms of Kullback-Leibler (KL) divergence. Specifically, the problem at hand is an optimisation challenge, framed as:
\begin{equation}
q^*(\mathbf{z})=\underset{q(\mathbf{z}) \in \mathcal{Q}}{\arg \min } \mathrm{KL}(q(\mathbf{z}) \| p(\mathbf{z} | \mathbf{x})) .
\end{equation}

Once found, $q^*(\cdot)$ is the best approximation of the conditional distribution within the chosen family \( \mathcal{Q} \). The computational complexity of this optimisation is dictated by the complexity of the family \( \mathcal{Q} \).

Nevertheless, minimising this objective is infeasible due to computing the evidence $\log p(\mathbf{x})$ in Eq. \ref{eq:marginal}. Recall that KL divergence is
\begin{equation}
\mathrm{KL}(q(\mathbf{z}) \| p(\mathbf{z} | \mathbf{x}))=\mathbb{E}[\log q(\mathbf{z})]-\mathbb{E}[\log p(\mathbf{z} | \mathbf{x})],
\end{equation}
where all expectations are taken with respect to $q(\mathbf{z})$. 
Upon expanding the conditional term, we see that the KL divergence is intimately tied to \( \log p(\mathbf{x}) \):
\begin{equation}
\mathrm{KL}(q(\mathbf{z}) \| p(\mathbf{z} | \mathbf{x}))=\mathbb{E}[\log q(\mathbf{z})]-\mathbb{E}[\log p(\mathbf{z}, \mathbf{x})]+\log p(\mathbf{x}) .
\label{kl}
\end{equation}

To circumvent this computational challenge, we optimise an alternative function known as the Evidence Lower Bound (ELBO):
\begin{equation}
\operatorname{ELBO}(q)=\mathbb{E}[\log p(\mathbf{z}, \mathbf{x})]-\mathbb{E}[\log q(\mathbf{z})] .
\end{equation}
The ELBO is the lower bound of the evidence $(\log p(x))$ after applying Jensen's inequality. 
Also, the ELBO is related to KL divergence as the ELBO is the negation of the KL divergence offset by the constant term \( \log p(\mathbf{x}) \).
Therefore, minimising the KL divergence will be the same as maximising the ELBO regardless of which distribution \(q\) we choose from the family \(\mathcal{Q}\). However,  when $q = p^{*}$, this resulting maximising ELBO will reduce the KL divergence to zero.

We can have intuitions about the optimal variational density by testing the ELBO.  Rewrite the ELBO as a sum of the expected log likelihood of the data and the KL divergence between the prior $p(\mathbf{z})$ and $q(\mathbf{z})$,
\begin{equation}
\begin{aligned}
\operatorname{ELBO}(q) & =\mathbb{E}[\log p(\mathbf{z})]+\mathbb{E}[\log p(\mathbf{x} | \mathbf{z})]-\mathbb{E}[\log q(\mathbf{z})] \\
& =\mathbb{E}[\log p(\mathbf{x} | \mathbf{z})]-\operatorname{KL}(q(\mathbf{z}) \| p(\mathbf{z}))
\end{aligned}
\end{equation}

In summary, maximising the ELBO not only offers a computationally tractable objective but also ensures that the selected \( q \) approximates the true posterior as closely as possible within the chosen distributional family.

\subsection{Rao Blackwellised particle filter}

The Rao-Blackwellized Particle Filters (RBPF) leverage the Rao-Blackwell theorem to enhance the sampling efficiency within particle filters by marginalising certain variables. While standard particle filters often falter in high-dimensional spaces due to the need for numerous particles, the integration of RBPFs significantly reduces the requisite particle count.

RBPF is aim to derive an estimator for the conditional distribution $p\left(y_t | z_t\right)$ such that less particles will be required to reach the same accuracy as  particle filter. The objective remains to estimate the distribution $p\left(z_t | y_{1: t}\right)$. The posterior can be expressed as:
$$
p\left(z_{0: t} | y_{1: t}\right)=p\left(z_{0: t-1} | y_{1: t-1}\right) \frac{p\left(y_t | z_t\right) p\left(z_t | z_{t-1}\right)}{p\left(y_t | y_{1: t-1}\right)}.
$$
Here sampling is used due to the difficulty of solving analytically .  Break the hidden variables $z$ into two groups: $r_t$ and $x_t$ such that $p\left(z_t | z_{t-1}\right)=p\left(x_t | r_{t-1: t}, x_{t-1}\right) p\left(r_t | r_{t-1}\right)$. By breaking them into two groups, they can determine part of the posterior $p\left(x_{0: t} | y_{1: t}, r_{0: t}\right)$ analytically followed by marginalise out the variables $x_{0: t}$. Subsequently, the posterior can be factorised as:
$$
p\left(r_{0: t}, x_{0: t} | y_{1: t}\right)=p\left(x_{0: t} | y_{1: t}, r_{0: t}\right) p\left(r_{0: t} | y_{1: t}\right).
$$
Given that the dimensions of \( p\left(r_{0: t} | y_{1: t}\right) \) are more compact than those of the original \( p\left(r_{0: t}, x_{0: t} | y_{1: t}\right) \), so they can obtain better results.

\section{S4 split}
\label{sec:S4_split}
The existing methods used for forecasting face difficulties when dealing with irregular time series data. This issue is particularly noticeable, in the S4 model. Time series data consists of data points arranged in an order and irregularities can occur due to missing data points, unexpected spikes or anomalies. These irregular patterns disrupt the consistency that models rely on for predictions.

While the S4 model is effective in ways it is especially susceptible to these inconsistencies. When confronted with irregular time series data its internal mechanisms may struggle to identify the underlying patterns or may mistake the irregularities as features. Consequently, its forecasting accuracy can significantly decrease. Therefore there is a need for strategies or improvements to enhance models, like S4 so that they can better handle the challenges presented by irregular time series data.

To address this problem,  we introduce a novel model named ``S4 split''. The S4 split is a version of its predecessor, the S4 and it specifically focuses on handling and analysing irregularities that can be found in time series data.

The unique mechanisms and features incorporated into the S4 split model allow it to distinguish between data patterns and irregularities. Unlike the S4 model, which may struggle when faced with inconsistencies the S4 split actively acknowledges and adjusts, for these discrepancies. This is made possible through algorithms and filtering processes integrated into the model.

Moreover, the design philosophy behind the S4 split revolves around segmenting and analysing data in parts. This ensures that even if one segment exhibits irregularities it won't have an impact on predictions. Adopting this approach it offers an adaptable method for forecasting time series data.

Initial tests and comparisons demonstrate that the S4 split performs better than the S4 model, in scenarios involving irregular time series data.

We utilise the Change Finder algorithm to identify change points, allowing us to segment the dataset based on these points. Next, using statistical approaches, we identify and amalgamate similar segments through linear interpolation. This process yields multiple distinct datasets, each reflecting unique statistical characteristics. We then match these new datasets with the most recent data to identify the relevant ones and subsequently train them using the S4 model. Fig. \ref{fig:S4 split} is the flow chart of the S4 split. 

\subsection{Split and merge datasets}

\begin{figure}[b]
    \begin{minipage}{\linewidth}
        \centering
        \includegraphics[width=1\linewidth]{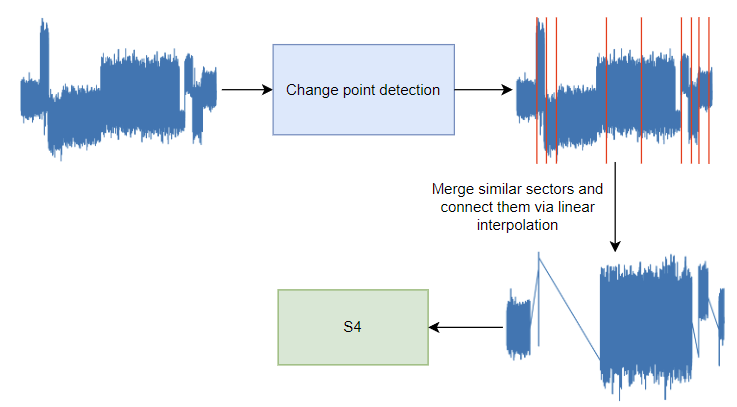}
        \caption{Flow chart of S4 split.}
        \label{fig:S4 split}
    \end{minipage}
    \vskip10pt
    \begin{minipage}{\linewidth}
        \centering
        \includegraphics[width=1\linewidth]{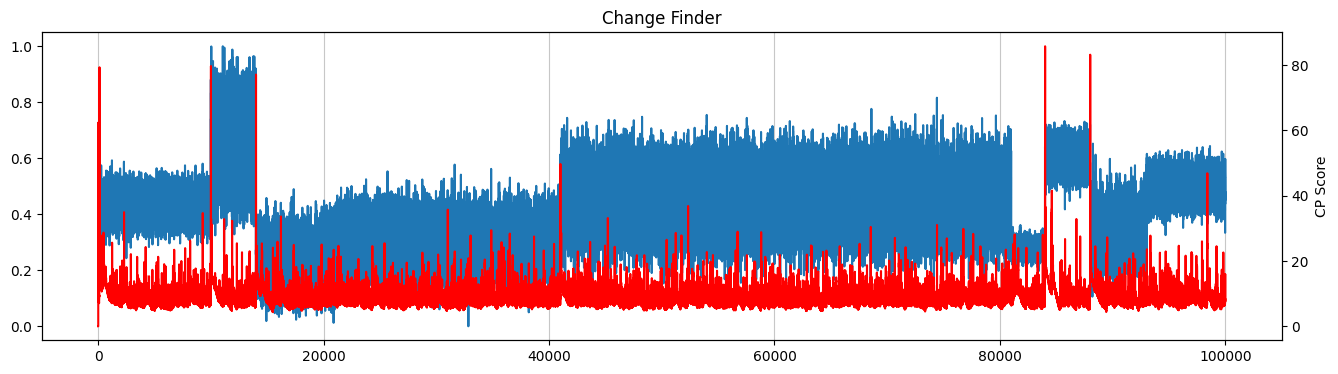}
        \caption{Result of change point detection.}
        \label{fig:changefinder}
    \end{minipage}
\end{figure}

To segment the data, we initiate by employing the Change Finder algorithm on the original dataset, yielding a change point $Score$. The outcome of this process on a random switching dataset is illustrated in Fig. \ref{fig:changefinder}. Evaluating the change point $Score$, we discern the change points, allowing for the dataset to be partitioned accordingly.

Feature extraction plays a role in understanding and analysing data segments. During this process, we focus on each segment individually. Extract statistical measures such as the average (mean) and the spread (standard deviation). These metrics are chosen because they provide information about the dataset's tendency and variability, giving us insights into underlying patterns or trends. The mean represents the value of the data points in a segment giving us an understanding of its central value. Conversely, the standard deviation gives us an idea of how the data points are spread out or tightly clustered around this mean. A higher standard deviation indicates variability, while a lower one suggests consistency. Once we have extracted these features, we meticulously analyse them to draw comparisons between different segments. By examining similarities and differences in their means and standard deviations, we can effectively categorise them based on their characteristics. Segments that exhibit features indicating behavioural patterns are then consolidated together for further analysis.

Our next step is to consolidate those similar segments. Since the datasets are time series, we cannot just stack them together as there will be a gap between each segment. To overcome this problem we use linear interpolation, which is a simple but efficient technique. This approach smoothly connects the end of one segment to the beginning of the subsequent segment ensuring that the chronological order is preserved. Also, applying linear interpolation can prevent no additional information will being captured by the S4 model. This is crucial when the data is subsequently processed by the S4 model, as it ensures that the model isn't thrown off or misinformed by any artificially introduced data points.

\subsection{S4}

After we split and merge datasets, we are ready to feed them into the S4 model. For this purpose, we employ the foundational S4 algorithm developed by Albert\cite{b17}. To bolster computational speed while maintaining accuracy, we incorporate techniques from S4D\cite{b30} and FlashConv\cite{b26} into the S4 framework.

To combine S4D and FlashConv technique into S4, we change the way that how S4 compute the convolution kernel $\overline{\mathbf{K}}$. More specifically we are using Vandermonde matrix-vector multiplication to compute $\overline{\mathbf{K}}$ and using Fused Block FFT convolution to compute $y$. Algorithm \ref{alg:s4} shows the detailed way to compute S4 layer.

\begin{algorithm}[b]
\caption{S4 layer}\label{alg:s4}
\begin{algorithmic}[1]
\State{Input: $ u \in \mathbb{R}^N, \mathrm{SSM} \text { parameterised by matrices } \mathbf{A} \in \mathbb{R}^{m \times m}, \mathbf{B} \in \mathbb{R}^{m \times 1}, \mathbf{C} \in \mathbb{R}^{1 \times m}, \mathbf{D} \in \mathbb{R}^{1 \times 1}, \text { chunk size } N^{\prime} \text { where } N \text { is a multiple of } N^{\prime} . $}
\State{Precompute $\mathbf{A} =\exp(\mathbf{A_{Re}}) + i \cdot \mathbf{A_{Im}}$ as Vandermonde matrix}
\State{Precompute $\mathbf{A}^{N^{\prime}} , \mathbf{M}_{u x}=[\mathbf{A}^{N^{\prime}-1} \mathbf{B}, \ldots, \mathbf{B}] \in \mathbb{R}^{m \times N^{\prime}}, \mathbf{M}_{x y}=[\mathbf{C}, \ldots, \mathbf{C A}^{N^{\prime}-1}] \in \mathbb{R}^{N^{\prime} \times m} .$}
\State{$\text {Split the inputs } u_{1: N} \text { into } C=N / N^{\prime} \operatorname{chunks} u_{1: N^{\prime}}^{(c)} \text { for } c=1, \ldots, C . $}
\State{$\text {Let the initial state be } x_{N^{\prime}}^{(0)}=0 \in \mathbb{R}^m .$}
\For{$1 \leq c \leq C$}
    \State{$\text { Compute } y^{(c)}=\mathbf{M}_{x y} x_{N^{\prime}}^{(c-1)}+\operatorname{Fused Block FFT convolution}\left(\overline{\mathbf{K}}, u_j\right)+\mathbf{D} u^{(c)} \in \mathbb{R}^{N^{\prime}} .$}
    \State{$\text {Update state: } x_{N^{\prime}}^{(c)}=\mathbf{A}^{N^{\prime}} x_{N^{\prime}}^{(c-1)}+\mathbf{M}_{u x} u^{(c)} .$}
\EndFor
\State{Return: $y=\left[y^{(1)} \ldots y^{(C)}\right]$} 
\end{algorithmic}
\end{algorithm}

To validate our approach, we juxtaposed our enhanced S4 with the standard S4, applying both to the LRA Tasks. The comparative results of this experimentation can be perused in Table \ref{tab:lra}. We can see that our S4 model are highly competitive on all datasets and achieved an average of near $85\%$ accuracy on the LRA task. Additionally, as illustrated in Table \ref{tab:time}, by leveraging FlashConv, we managed to accelerate S4 performance by $2 \times$, achieving a significant edge over Transformers with a speed improvement of $5.8 \times$.

\begin{table}[t]
    \centering
    \caption{Performance on Long Range Arena (LRA) Tasks. ${*}$ AAN task is not included due to computer issues. }
    \begin{adjustbox}{width=1\linewidth}
    \begin{tabular}{lcccccc}
    \toprule
        Model & ListOps & IMDB 	& CIFAR	& Pathfinder & Path-X & Avg. \\
\midrule
S4 (original)\cite{b17}   & 59.5	& 86.5 &	88.5 &	94.0 &	96.0  &  84.9\\
S4 modified  & 59.3	& 86.7  &	89.5 &	93.6 &	95.4  & 84.7 \\
\bottomrule
\end{tabular}
\end{adjustbox}
\label{tab:lra}
\end{table}

\begin{table}[b] 
    \centering
    \caption{Speed up on the LRA benchmark.}
    \begin{adjustbox}{width=0.4\linewidth}
    \begin{tabular}{lcccccc|c}
    \toprule
        Model & Speed up \\
\midrule
   Transformer \cite{b4} & $1 \times$ \\
 S4 \cite{b17}  & $2.9 \times$ \\
 S4 modified  & $5.8 \times$ \\
\bottomrule
\end{tabular}
\end{adjustbox}
\label{tab:time}
\end{table}

For testing the S4 model in LRA task, we use the same hyperparameters setting for the S4 model in \cite{b17}. Hyperparameters for all datatsets are reported in Table \ref{tab:hyperparameters}. For all the test, we use AdamW optimiser with a constant learning rate. The S4 state size is always fixed to $N=64$. Since LRA tasks are classification, so we add mean pooling after the last layer. All the result were run on 3 A6000 GPU, which has 40 Gb memory.

\begin{table}[b] 
    \centering
    \caption{The values of the hyperparameters used on LRA Tasks. BN and LN refer to Batch Normalisation and Layer Normalisation.\cite{b17}}
    \renewcommand{\arraystretch}{1.5}
    \begin{adjustbox}{width=1\linewidth}
    {\LARGE
    \begin{tabular}{lcccccccc}
    \toprule
         & Depth & Features & Norm & Dropout & Learning Rate & Batch Size & Epochs & Weight Decay\\
\midrule
ListOps   & 6	& 128 &	BN &	0 &	0.01  &  100 & 50 & 0.01\\
IMDB  & 4	&  64  &	BN &	0 &	 0.001  & 50  &20 &0\\
CIFAR  & 6	& 512  &	LN &	0.2 &	0.004  & 50 &200&0.01\\
Pathfinder  & 6	& 256  &	BN &	0.1 &	0.004  & 100&200 &0\\
Path-X  & 6	& 256  &	BN &0 &	 0.0005  & 32  &100 &0\\
\bottomrule
\end{tabular}
}
\end{adjustbox}
\label{tab:hyperparameters}
\end{table}

\begin{figure}
    \centering
    \includegraphics[width=1\linewidth]{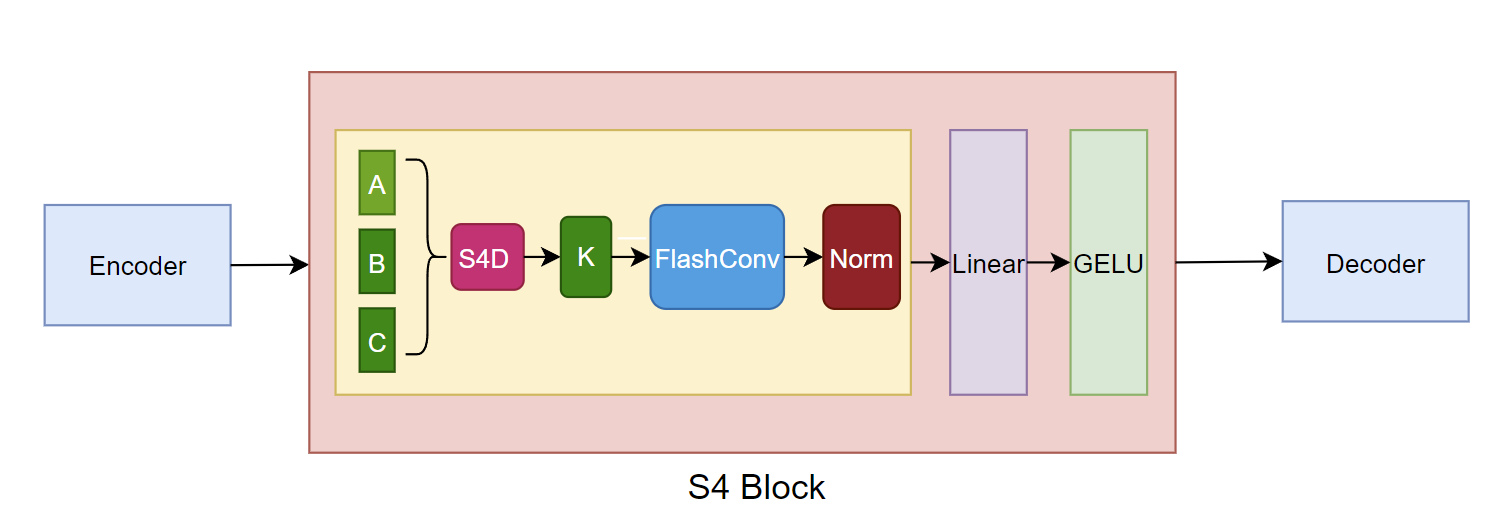}
    \caption{Structure of S4 model.}
    \label{fig:s4model}
\end{figure}

The S4 model structure here is simple, Fig. \ref{fig:s4model} shows the S4 structure. This architecture consists of a linear encoder to encode the input at each time step into $H$ features, a S4 block that contain multiple S4 layer and a linear decoder to decode the output of S4 block at each time step back to the input size.

Mean pooling layer and Softmax operation are only used for the classification tasks. The mean pooling layer compresses the output of the last S4 layer across the sequence length dimension, and a single $H$-dimensional encoding for softmax classification.

To initialise S4, we recall the HiPPO matrix $\mathbf{A}$ S4-Inv (\ref{s4d-inv}), the correspond matrix  $\mathbf{B} = \mathbf{1}$ and $\mathbf{C}$ initialise randomly with standard deviation 1 as the initial parameter to feed in.

\subsection{Experiment}
We apply our S4 split model on the simple 1-D switching time series as shown in Fig. \ref{fig:changefinder}. The sequence length is 100k with different mean and standard deviation normal distributions. For both reconstruction and prediction task, all the models' hyperparameters are shown in Table \ref{tab:hyperparameterss4}. All the models uses AdamW optimiser with constant learning rate.

\begin{table}[b] 
    \centering
    \caption{The values of the hyperparameters for testing LSTM, S4 and S4 split model.  LN refer to Layer Normalisation.}
    \renewcommand{\arraystretch}{1.5}
    \begin{adjustbox}{width=1\linewidth}
    {\LARGE
    \begin{tabular}{lccccccc}
    \toprule
         & Depth  	& Norm & Dropout & Learning Rate & Batch Size & Epochs & Weight Decay\\
         \midrule
        |rule
LSTM   & 1	 &	LN &0 &	0.001  &  200 & 3000 &0\\
S4  & 3	  &	LN &	0 &	 0.001  & 200  & 10000 &0\\
S4 split  & 3	  &	LN &	0 &	0.001  & 200 &10000&0\\
\bottomrule
\end{tabular}
}
\end{adjustbox}
\label{tab:hyperparameterss4}
\end{table}

\subsubsection{Reconstruction}

\begin{figure}
    \centering
    \includegraphics[width=1\linewidth]{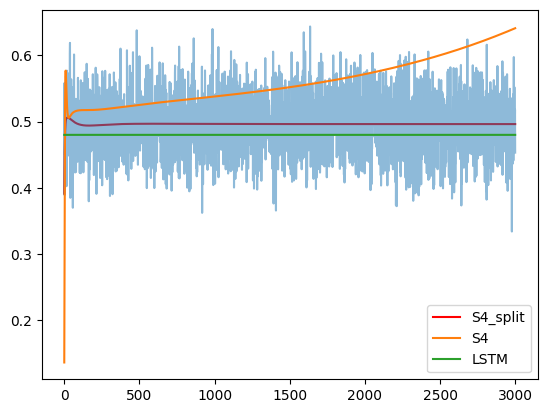}
    \caption{Reconstruction on 1-D switching time series.}
    \label{fig:enter-rec}
\end{figure}

We initially focus on a reconstruction task involving a 1-D switching time series. For this task, we train the model on 100k data points with feed in a length of 100k random Gaussian distribution for the input data. The results, which can be observed in Fig. \ref{fig:enter-rec} and Table \ref{tab:Reconstruction}, demonstrate the effectiveness of S4 Split in handling the switching time series. Notably, S4 Split outperforms the original S4 model by $5 \times$. This performance gap becomes evident as the pattern generated by S4 diverges significantly from the original data towards the end of the sequence. This also shows that vanilla S4 model does not perform well when the state of $x$ is changing. Furthermore, we have performed a prediction task to further prove our result.

\begin{table}[b] 
    \centering
    \caption{Reconstruction on 1-D switching time series.}
    \begin{adjustbox}{width=0.27\linewidth}
    \begin{tabular}{lc}
    \toprule
        Model & MSE \\
\hline
   LSTM \cite{b4} &  0.007 \\
 S4 \cite{b17}  & 0.023 \\
 S4 split  & 0.004 \\
\bottomrule
\end{tabular}
\end{adjustbox}
\label{tab:Reconstruction}
\end{table}

\subsubsection{Prediction}

We performed prediction task on 1-D switching time series. We trained the model on 100k data points, and predict the next 3k data. The prediction methods for S4 is different to the LSTM.

For LSTM, we use the RNN architecture to predict the subsequent data point, and add this points into the original dataset as a new dataset, then use the new dataset to forecast the following data point, and this process is iteratively repeated to continue making predictions. 

In contrast to LSTM, both S4 and S4 split are using CNN architecture. The goal is to forecast a continuous range of future data $F$ based on a range of past data $P$. To accomplish this, we concatenate $P$ and $F$ together and using mask technique to mask the future data, i.e., $P \times 1 + F \times 0$. Since we are using convolution kernel,  described by $y = \overline{\mathbf{K}} * u$, if $u = 0$ implies $y=0$, which not effect the result. Hence when we training the model, we will mask $F$ parts. This will input a sequence of length $P+F$ and run through S4 model, and results in an output of the same length $P+F$, where $F$ segment serves as our forecast.

\begin{figure}[b]
    \centering
    \includegraphics[width=1\linewidth]{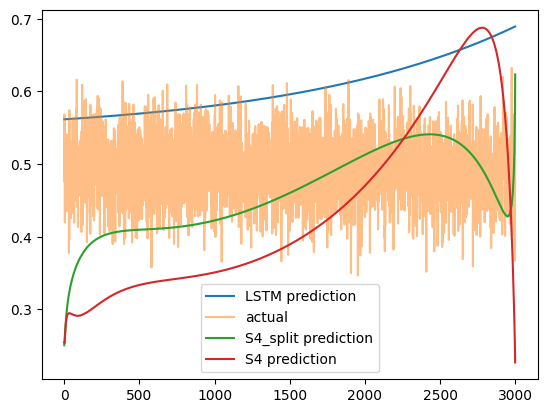}
    \caption{Prediction on 1-D switching time series.}
    \label{fig:enter-pre}
\end{figure}

The effectiveness of S4 Split in handling switching time series is further substantiated by the results presented in Fig. \ref{fig:enter-pre} and Table \ref{tab:prediction}. S4 Split outperforms the original S4 model by a factor of 4 and surpasses LSTM by a factor of 3.5. Additionally, these results highlight a distinct advantage of using S4: its capacity to handle long-term dependencies. In particular, S4 split and S4 demonstrates superior performance when predicting as far as 10k steps ahead, a feat that LSTM is unable to achieve.

\begin{table}[t] 
    \centering
    \caption{Prediction on 1-D switching time series.}
    \begin{adjustbox}{width=0.5\linewidth}
    \begin{tabular}{lcc}
    \toprule
        Model & MSE (3k)&  MSE (10k)\\ 
    \midrule
LSTM \cite{b4} &  0.027 & 0.17\\
 S4 \cite{b17}  & 0.031& 0.098 \\
 S4 split  & 0.008 & 0.024\\
\bottomrule
\end{tabular}
\end{adjustbox}
\label{tab:prediction}
\end{table}

However, This methods only works in offline setting where there will be no change point happens in the prediction part, which is not working in real world data sets like weather forecasting. Hence we need to find another methods that can solve this problem. Therefore we are going to present our new model S4\_SNLDS.

\section{S4 SNLDS }
\label{S4_SNLDS}

A Switching Linear Dynamical System model captures a complex system's non-linear dynamic behaviour by transitioning between a suite of either linear dynamic models or non-linear dynamic models over time. For example, an aeroplane flying across the country. The movement of the aircraft consists of linear dynamics and non-linear dynamics. When the fight path is changed from one dynamic to the other, the switch can depend on the system's discrete state or external factors. By identifying these behavioural segments and understanding the underlying switching mechanisms, we can delve deeper into the intricate patterns that bring about complex natural events. Many models are trying to solve such problems. For example, Linderman\cite{b29} introduces a class of recurrent state space models adept at encapsulating these inherent relationships. Accompanying this, he presents Bayesian inference and learning techniques that are not only computationally tractable but also scalable to large datasets. Also, Zhe\cite{b22} proposes a variational inference algorithm for fitting SNLDS adept at segmenting high-dimensional signal time series into distinct, meaningful temporal segments or ``modes''. Nevertheless, current approaches do not perform well in foresting time series that contain long-range dependencies, and the forecasting accuracy decreases dramatically as the time step increases (detail in Section \ref{Lorenz Attractor}). Moreover, Both models use RNN architecture with a limited memory horizon, which will lose information from the early stage.  Therefore there is a need for improvements to enhance models, like SNLDS so that they can better handle the challenges presented by long term dependency.

To address this problem, we introduce a novel model, S4 SNLDS, that combines the strength of both S4 and SNLDS, the long-range dependency ability, identifies the switch behaviour and learns the underlying relation between the discrete state and stitches. Moreover, S4 SNLDS adapted the all the benefit from S4 split that can do everything S4 split do. 

Lorenz Attractor test and 2-D bouncing ball test demonstrate that S4 SNLDS outperforms SNLDS in both reconstruction and prediction task.

\subsection{Model Formulation}

We use the same state space model as we defined earlier in Section \ref{sec:snlds} as our generative model. The model consist three part:
\begin{itemize}
    \item Emission distribution: $p(\mathbf{x}_t|\mathbf{z}_t)$ shows how to generate the observation by the given continues hidden states.
    \item Continues state transition: $p(\mathbf{z}_t|\mathbf{z}_{t-1},s_t)$ shows how to find the continues dynamics transit by the given the discrete mode.
    \item Discrete state transition: $p(s_t|s_{t-1},\mathbf{x}_{t-1})$ shows the discrete state transition is either depends on the hidden dynamics or the observations. 
\end{itemize}

The functional form of the conditional probability distributions for S4 SNLDS are defined as:
\begin{equation}
\begin{aligned}
p\left(\mathbf{x}_t | \mathbf{z}_t\right) &= \mathcal{N}\left(\mathbf{x}_t | {S4}_x\left(\mathbf{z}_t\right), \mathbf{R}\right), \\
p\left(\mathbf{z}_t | \mathbf{z}_{t-1}, s_t=k\right) &= \mathcal{N}\left(\mathbf{z}_t | {S4}_z\left(\mathbf{z}_{t-1}, k\right), \mathbf{Q}\right), \\
p\left(s_t | s_{t-1}=j, \mathbf{x}_{t-1}\right) &= \operatorname{Cat}\left(s_t | \mathcal{S}\left(f_s\left(\mathbf{x}_{t-1}, j\right)\right)\right),
\end{aligned}
\end{equation}
where \( f_{s} \) are non-linear functions, ${S4}_x , {S4}_z$ are S4 blocks which are non-linear functions as well.

Now, by leveraging RBPF, we can analytically marginalising out some of latent variables. We can conditioned on knowing $\mathbf{z}_{1:T}$ and $\mathbf{x}_{1:T}$, we can marginalise out $s_{1:T}$ in linear time via the forward-backward algorithm, so we can handle non-linear models. However, it is difficult to compute the optimal distribution $p(\mathbf{z}_t|\mathbf{z}_{t-1},\mathbf{x})$, therefore we can use variational inference to approximate this.

\subsection{Inference}

Define the variational posterior: 
$$q_{\phi, \boldsymbol{\theta}}(\mathbf{z}, \mathbf{s} | \mathbf{x})= q_{\boldsymbol{\phi}}(\mathbf{z} | \mathbf{x}) p_{\boldsymbol{\theta}}(\mathbf{s} | \mathbf{z}, \mathbf{x}),$$
where $p_{\boldsymbol{\theta}}(\mathbf{s} | \mathbf{z}, \mathbf{x})$ is the exact posterior computed using the forward-backward algorithm. 
To compute $q_\phi(\mathbf{z} | \mathbf{x})$, we first process $\mathbf{x}_{1: T}$ through a RNN mode S4 that HiPPO matrix $\mathbf{A}$ can extract features from $\mathbf{x}_{1:T}$ to the states $\mathbf{h}_t^x$. It can be described as:
\begin{equation}
\begin{aligned}
    \mathbf{h}_{t+1}^x &= \mathbf{A}^x \mathbf{h}_{t}^x + \mathbf{B}^x \mathbf{v}_t, \mathbf{v}_t \stackrel{\mathrm{iid}}{\sim} \mathcal{N}\left(0, 1 \right),  \\
    \mathbf{x}_t& = \mathbf{C}^x \mathbf{h}_{t}^x. 
\end{aligned}
\end{equation}

Although this is a linear transformation, we will apply a non-linear activation function to the output; hence, it still is a non-linear function.
Similarly, we apply another RNN mode S4 to compute the parameters of $q\left(\mathbf{z}_t | \mathbf{z}_{1: t-1}, \mathbf{x}_{1: T}\right)$ with the state $\mathbf{h}_t^z$,  where the hidden state is computed based on $\mathbf{h}_{t-1}^z$ and $\mathbf{h}_t^x$, i.e., $\mathbf{h}_t^z = \mathbf{A}^z \mathbf{h}_{t-1}^z + \mathbf{B}^z \mathbf{h}_t^x$. This gives the following approximate posterior: $q_\phi\left(\mathbf{z}_{1: T} | \mathbf{x}_{1: T}\right)=$ $\prod_t q\left(\mathbf{z}_t | \mathbf{z}_{1: t-1}, \mathbf{x}_{1: T}\right)=\prod_t q\left(\mathbf{z}_t | \mathbf{h}_t^z\right)$. 

\begin{figure}[t]
    \centering
    \includegraphics[width=1\linewidth]{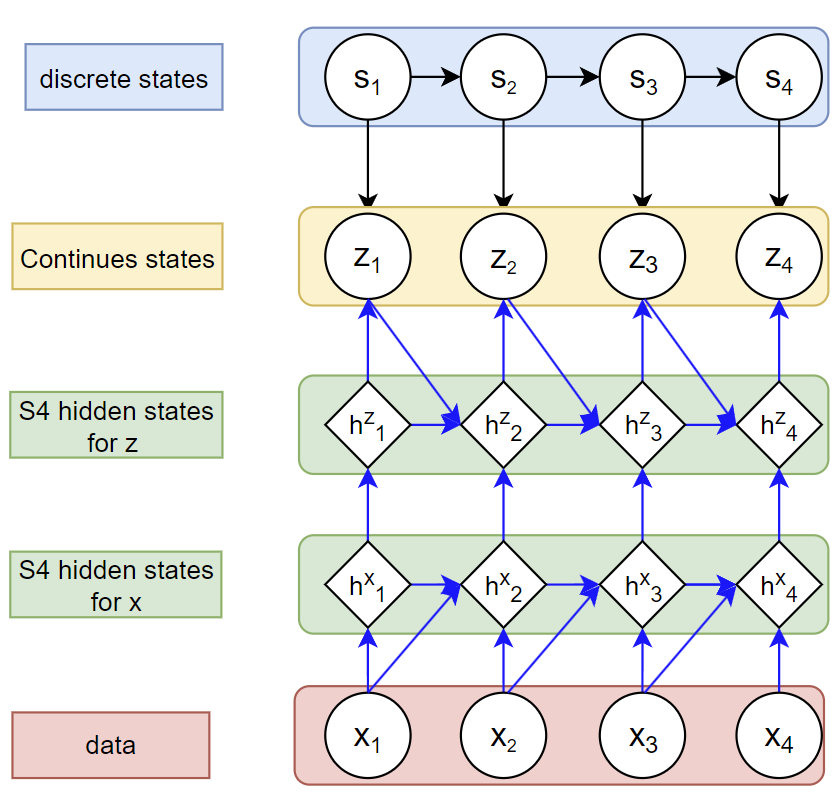}
    \caption{Inference network for the S4 SNLDS. Black arrows share parameters $\theta$ for generative model. Blue arrows have parameters $\phi$ for $q_{\phi}(z|x)$. The diamonds represents deterministic nodes computed with S4s. $h_t^x$ is a S4 applied to $\mathbf{x}_{1:T}$ and $h_t^z$ is a S4 applied to $\mathbf{h}^x_{t-1}$ and $\mathbf{z}_{t-1} $.}
    \label{fig:s4snlds}
\end{figure}

Fig. \ref{fig:s4snlds} shows the illustration of the inference network.

\subsection{Learning}

The evidence lower bound objective (ELBO) of the model can be expressed as:
\begin{equation}
\begin{aligned}
\mathcal{L}_{ELBO}= & \mathbb{E}_{q_{\theta, \phi}(\mathbf{z}, \mathbf{s} | \mathbf{x})}\left[\log p_\theta(\mathbf{x}, \mathbf{z}, \mathbf{s})-\log q_{\theta, \phi}(\mathbf{z}, \mathbf{s} | \mathbf{x})\right] \\
= & \mathbb{E}_{q_\phi(\mathbf{z} | \mathbf{x}) p_\theta(\mathbf{s} | \mathbf{x}, \mathbf{z})}\left[\log p_\theta(\mathbf{x}, \mathbf{z}) p_\theta(\mathbf{s} | \mathbf{x}, \mathbf{z})\right. \\
  & \left.- \log q_\phi(\mathbf{z} | \mathbf{x}) p_\theta(\mathbf{s} | \mathbf{x}, \mathbf{z})\right] \\
= & \mathbb{E}_{q_\phi(\mathbf{z} | \mathbf{x})}\left[\log p_\theta(\mathbf{x}, \mathbf{z})\right]+H\left(q_\phi(\mathbf{z} | \mathbf{x})\right)
\label{elbo}
\end{aligned}
\end{equation}
where the $\mathbb{E}_{q_\phi(\mathbf{z} | \mathbf{x})}\left[\log p_\theta(\mathbf{x}, \mathbf{z})\right]$ is the model likelihood, and $H\left(q_\phi(\mathbf{z} | \mathbf{x})\right)$ is the conditional entropy for variational posterior associated with continuous hidden states.

To approximate the entropy of \(q_\phi(\mathbf{z} | \mathbf{x})\), we have:
$$
H\left(q_\phi(\mathbf{z} | \mathbf{x})\right)=H\left(q_\phi\left(\mathbf{z}_1\right)\right)+\sum_{t=2}^T H\left(q_\phi\left(\mathbf{z}_t | \tilde{\mathbf{z}}_{1: t-1}\right)\right)
$$
where $\tilde{\mathbf{z}}_t \sim q\left(\mathbf{z}_t\right)$ is a sample from the variational posterior, which means the gradient can be computed by the backpropagtion from the inference RNN.

The gradient for $\mathbb{E}_{q_\phi(\mathbf{z} | \mathbf{x})}\left[\log p_\theta(\mathbf{x}, \mathbf{z})\right]$ wrt $\theta$ can be approximated as:
$$
\nabla_\theta \mathbb{E}_{q_\phi(\mathbf{z} | \mathbf{x})}\left[\log p_\theta(\mathbf{x}, \mathbf{z})\right]=\mathbb{E}_{q_\phi(\mathbf{z} | \mathbf{x})}\left[\nabla_\theta \log p_\theta(\mathbf{x}, \mathbf{z})\right]
$$
Then apply reparameterisation trick for the gradient wrt $\phi$:
$$
 \nabla_\phi \mathbb{E}_{q_\phi(\mathbf{z} | \mathbf{x})}\left[\log p_\theta(\mathbf{x}, \mathbf{z})\right] 
 =\mathbb{E}_{\boldsymbol{\epsilon} \sim N}\left[\nabla_\phi \log p_\theta\left(\mathbf{x}, \mathbf{z}_\phi(\boldsymbol{\epsilon}, \mathbf{x})\right)\right]
$$
Therefore, the gradient is expressed as:
$$
\begin{aligned}
\nabla_\theta \mathcal{L} & =\mathbb{E}_{q_\phi(\mathbf{z} | \mathbf{x})}\left[\nabla_\theta \log p_\theta(\mathbf{x}, \mathbf{z})\right], \\
\nabla_\phi \mathcal{L} & =\mathbb{E}_{\boldsymbol{\epsilon} \sim N}\left[\nabla_\phi \log p_\theta\left(\mathbf{x}, \mathbf{z}_\phi(\boldsymbol{\epsilon}, \mathbf{x})\right)\right]+\nabla_\phi H\left(q_\phi(\mathbf{z} | \mathbf{x})\right) .
\end{aligned}
$$
Define $\mathbf{w}=\left(\mathbf{x}_{1: T}, \mathbf{z}_{1: T}\right)$. Therefore
$$
\begin{aligned}
\nabla \log p(\mathbf{w})= & \mathbb{E}_{p(\mathbf{s} | \mathbf{w})}[\nabla \log p(\mathbf{w})], \quad(\log p(\mathbf{w})=\log p(\mathbf{w}, \mathbf{s})-\\& \log p(\mathbf{s} | \mathbf{w})) \\
= & \mathbb{E}_{p(\mathbf{s} | \mathbf{w})}[\nabla \log p(\mathbf{w}, \mathbf{s})]  -\mathbb{E}_{p(\mathbf{s} | \mathbf{w})}[\nabla \log p(\mathbf{s} | \mathbf{w})] \\
= & \mathbb{E}_{p(\mathbf{s} | \mathbf{w})}[\nabla \log p(\mathbf{w}, \mathbf{s})]-0
\end{aligned}
$$
since
$$
\begin{aligned}
\mathbb{E}_{p(\mathbf{s} | \mathbf{w})}[\nabla \log p(\mathbf{s} | \mathbf{w})]  =\int p(\mathbf{s} | \mathbf{w}) \frac{\nabla p(\mathbf{s} | \mathbf{w})}{p(\mathbf{s} | \mathbf{w})} =\nabla 1=0 .
\end{aligned}
$$
Apply Markov property to $\nabla \log p(\mathbf{w}, \mathbf{s})$, we have
$$
\begin{aligned}
\nabla \log p(\mathbf{w}, \mathbf{s})= & \sum_{t=2}^T \nabla \log p\left(\mathbf{x}_t | \mathbf{z}_t\right) p\left(\mathbf{z}_t | \mathbf{z}_{t-1}, s_t\right) p\left(s_t | s_{t-1}, \mathbf{x}_{t-1}\right)  +\\& \nabla \log p\left(\mathbf{x}_1 | \mathbf{z}_1\right) p\left(\mathbf{z}_1 | s_1\right) p\left(s_1\right).
\end{aligned}
$$
Then take the expectation:

\begin{strip}
\begin{equation}
\begin{aligned}
\nabla \log p(\mathbf{w}) =& \mathbb{E}_{p(\mathbf{s} | \mathbf{v})}[\nabla \log p(\mathbf{w}, \mathbf{s})] \\
=& \sum_k p\left(s_1=k | \mathbf{w}\right) \nabla \log p\left(\mathbf{x}_1 | z_1\right) p\left(\mathbf{z}_1 | s_1=k\right) p\left(s_1=k\right) \\
& +\sum_{t=2}^T \sum_{j, k} p\left(s_{t-1}=j, s_t=k | \mathbf{w}\right) \nabla\left[\log p\left(\mathbf{x}_t | \mathbf{z}_t\right)  p\left(\mathbf{z}_t | \mathbf{z}_{t-1}, s_t=k\right) p\left(s_t=k | s_{t-1}=j, \mathbf{x}_{t-1}\right)\right] 
\end{aligned}
\label{elbotrain}
\end{equation}
\end{strip}

\subsection{Entropy regularisation}

From prior work \cite{b22}, they found out that SNLDS used  single discrete state, when nonlinear function is applied to model $p\left(\mathbf{z}_t | \mathbf{z}_{t-1}, s_t\right)$. To solve this problem, they add an regularising term to the ELBO that penalises the KL divergence between the state posterior at each time step and a uniform prior $p_{\text {prior }}\left(s_t=k\right)=1 / K$ . This is called a cross-entropy regulariser:
$$
\mathcal{L}_{\mathrm{CE}}=\sum_{t=1}^T \mathbb{K} \mathbb{L}\left(p_{\text {prior }}\left(s_t\right) \| p\left(s_t | \mathbf{z}_{1: T}, \mathbf{x}_{1: T}\right)\right)
$$
Hence the overall loss functions is
$$
\mathcal{L}(\boldsymbol{\theta}, \boldsymbol{\phi})=\mathcal{L}_{\mathrm{ELBO}}(\boldsymbol{\theta}, \boldsymbol{\phi})-\eta \mathcal{L}_{\mathrm{CE}}(\boldsymbol{\theta}, \boldsymbol{\phi}),
$$
where $\eta>0$ is a scaling factor. Set $\eta$ to be large at the beginning to ensure that all states are visited.

\subsection{Lorenz Attractor}
\label{Lorenz Attractor}

SLDS serve as a practical approximation for complex nonlinear dynamical systems. One of their primary advantages is that, once calibrated, they allow us to benefit from extensive research on optimal filtering, smoothing, and control methodologies specifically designed for linear systems. However, as we demonstrate using the example of the Lorenz attractor, traditional SLDS often fall short as generative models, particularly in their ability to interpolate over missing data points. 

To address this limitation, the S4 SNLDS enhances the standard model by bridging discrete ,continuous states and output. This modification improves the system's capability to handle data gaps, offering a more robust and reliable modelling framework.

The nonlinear dynamics Lorenz attractor are given by,
$$
\frac{\mathrm{d} \boldsymbol{x}}{\mathrm{d} t}=\left[\begin{array}{c}
\alpha\left(x_2-x_1\right) \\
x_1\left(\beta-x_3\right)-x_2 \\
x_1 x_2-\gamma x_3
\end{array}\right].
$$

We apply our S4 SNLDS model on the 1-D nonlinear dynamics Lorenz attractor data. We sample 500 steps for the sequence. For reconstruction and Generation task, the S4 SNLDS architecture contains an linear encoder, an inference RNN that contain two RNN S4 with both contain 3 S4 layer, both emission network and discrete state transition network are a 2-layer MLPs, the continuous state transition network used GRU followed by linear transformation and a linear decoder. And the learning rate is constant $10^{-4}$.

\subsubsection{Reconstruction}

\begin{figure}
    \centering
    \includegraphics[width=1\linewidth]{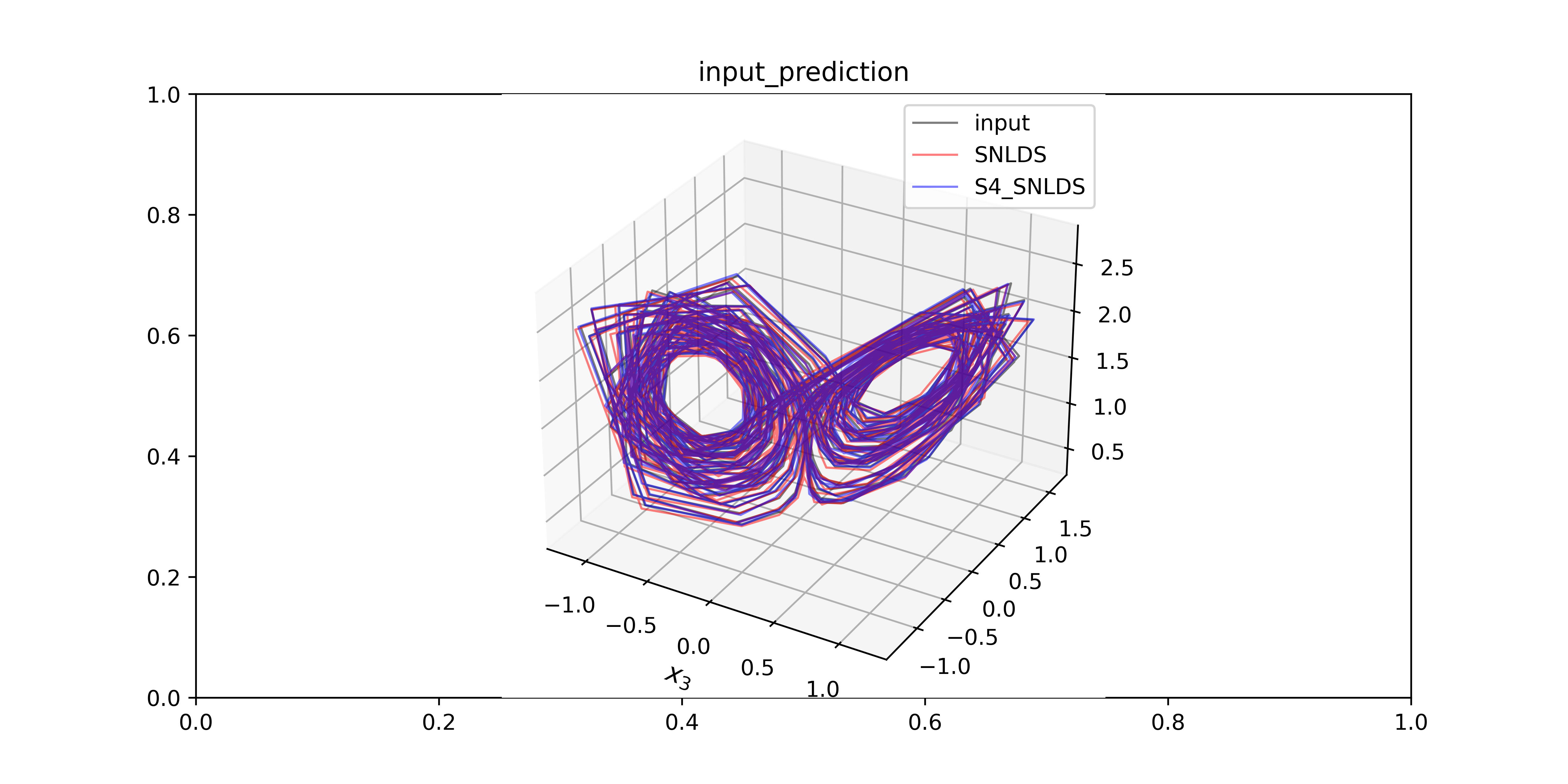}
    \caption{Lorenz attractor Reconstruction whole.}
    \label{fig:show_lorenz_attractor_rec}
\end{figure}

\begin{figure}
    \centering
    \includegraphics[width=1\linewidth]{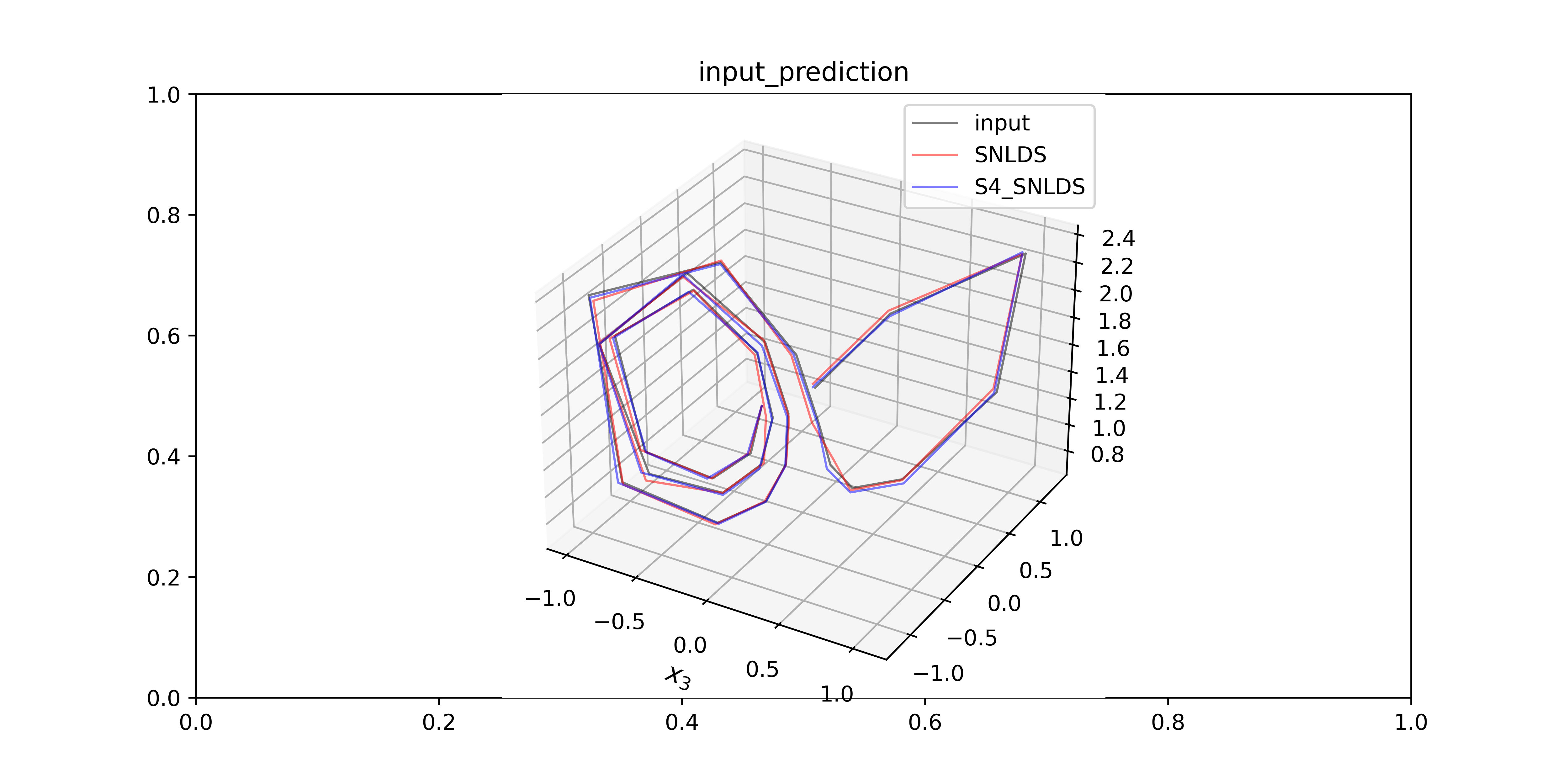}
    \caption{Lorenz attractor Reconstruction last 30 steps.}
    \label{fig:show_lorenz_attractor_3d_rec30}
\end{figure}

We trained both models with 500 data points with feed in a length of 500 random Gaussian distribution data to generate the 500 data points for the 1-D nonlinear dynamics Lorenz attractor. The result can be observed in Fig. \ref{fig:show_lorenz_attractor_rec}, Fig. \ref{fig:show_lorenz_attractor_3d_rec30} and Table \ref{tab:Reconstruction_snlds}, demonstrate the effectiveness of S4 SNLDS in handling both the long term dependency and the switching time series. Moreover, S4 SNLDS outperforms the SNLDS model by $3 \times$, and if we look Fig. \ref{fig:show_lorenz_attractor_3d_rec30} closely, we can see SNLDS is cannot detect the changes or delay the changes which shows how well S4 SNLDS performs. 

\begin{table}[b]
    \centering
    \caption{Reconstruction on 1-D switching time series.}
    \begin{adjustbox}{width=0.3\linewidth}
    \begin{tabular}{lc}
    \toprule
        Model & MSE \\
    \midrule
  SNLDS \cite{b22} &  0.008 \\
 S4 SNLDS   & 0.003 \\
\bottomrule
\end{tabular}
\end{adjustbox}
\label{tab:Reconstruction_snlds}
\end{table}

\subsubsection{Generation}

\begin{figure}
    \centering
    \includegraphics[width=1\linewidth]{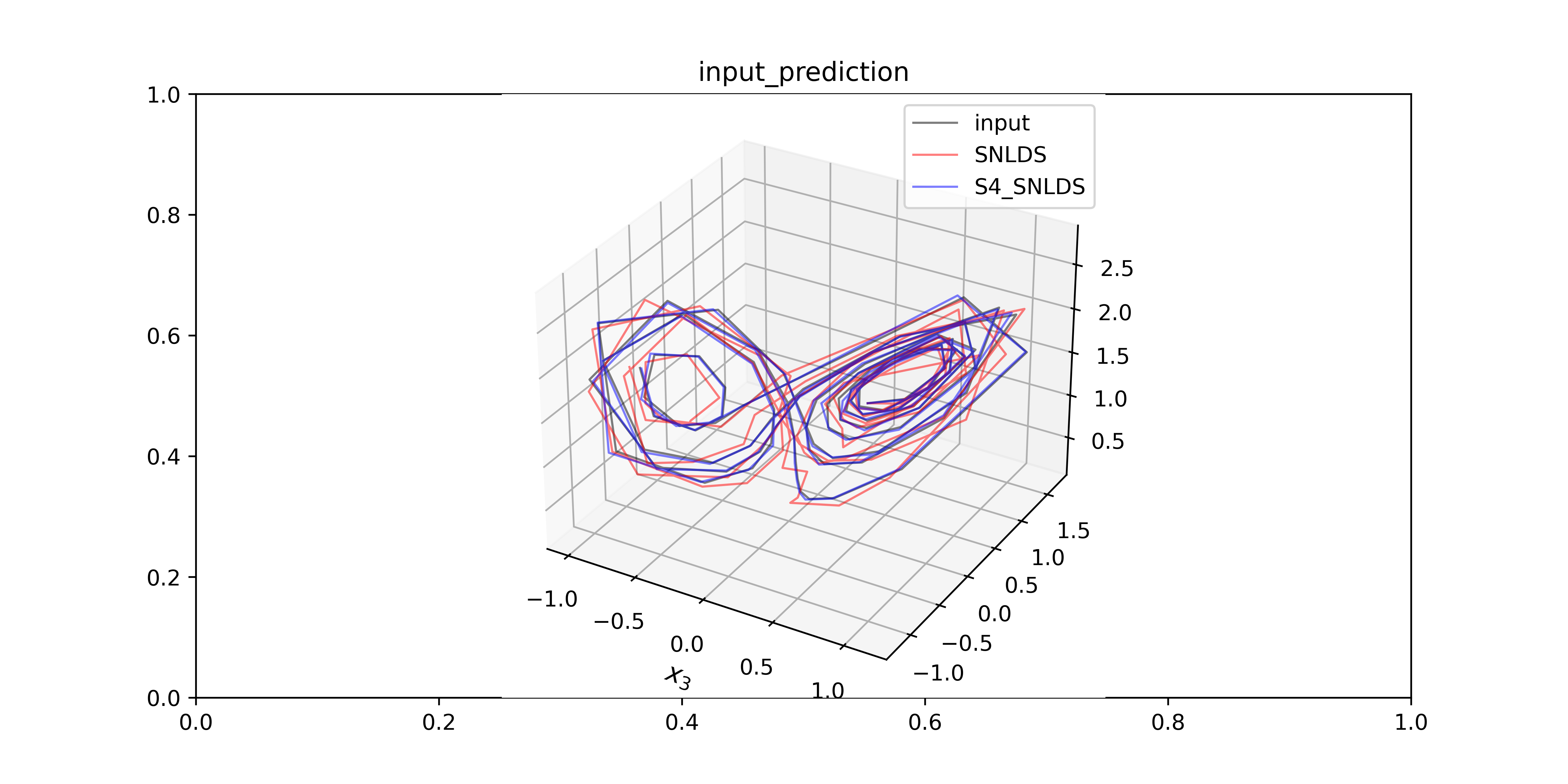}
    \caption{Lorenz attractor Reconstruction whole.}
    \label{fig:show_lorenz_attractor_100}
\end{figure}

\begin{figure}
    \centering
    \includegraphics[width=1\linewidth]{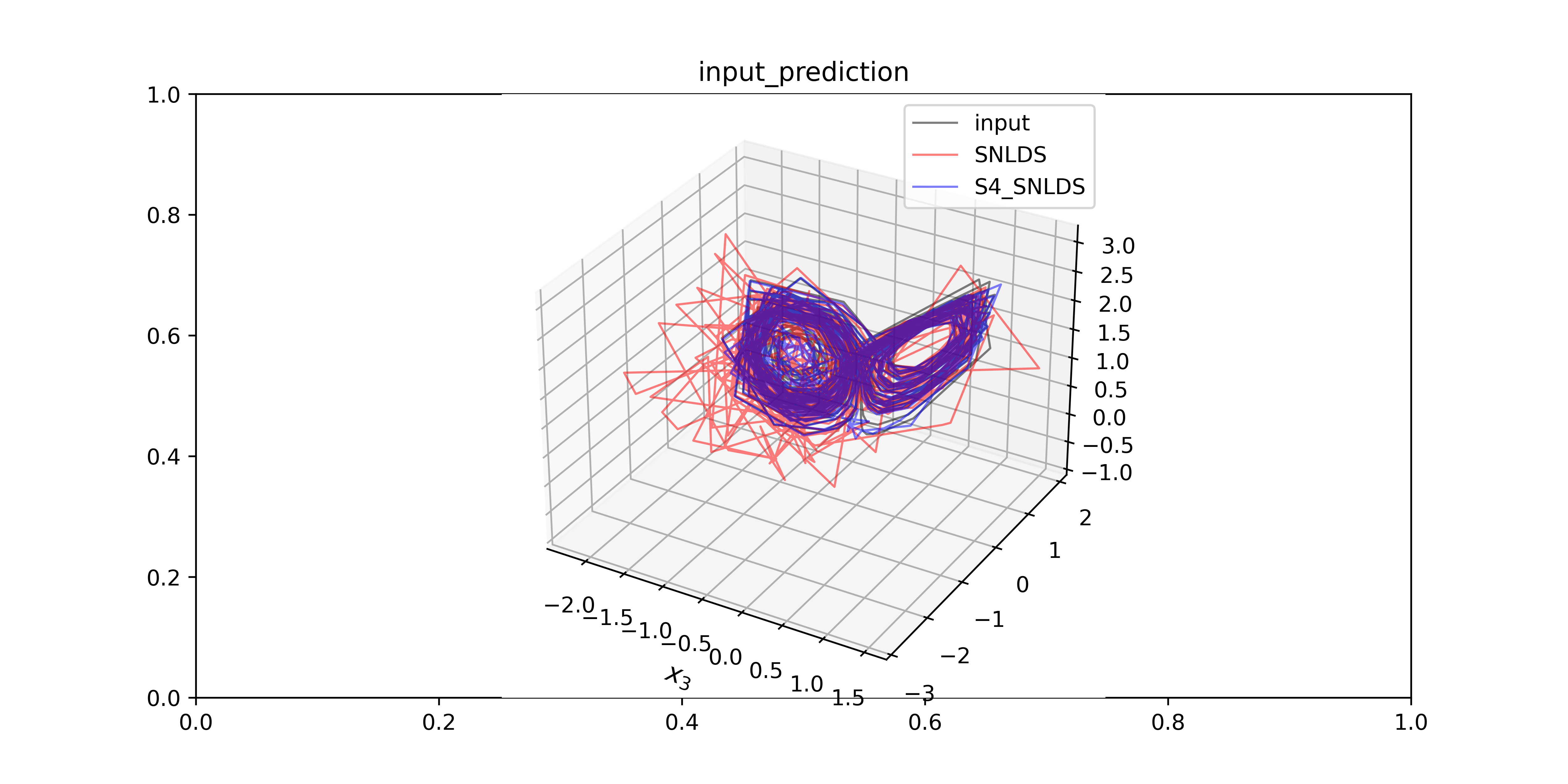}
    \caption{Lorenz attractor Reconstruction.}
    \label{fig:show_lorenz_attractor_500}
\end{figure}

We next explore the Generation task for the 1-D nonlinear Lorenz attractor dynamics. Using the 500 data points mentioned earlier, we initiate the generation process to evaluate the performance of both models, especially in terms of long-range dependencies. After generating an additional 100 steps, the results are depicted in Fig. \ref{fig:show_lorenz_attractor_100}. For a deeper dive, Fig. \ref{fig:show_lorenz_attractor_500} presents the outcomes after a 500-step generation. 

\begin{figure}
    \centering
    \includegraphics[width=1\linewidth]{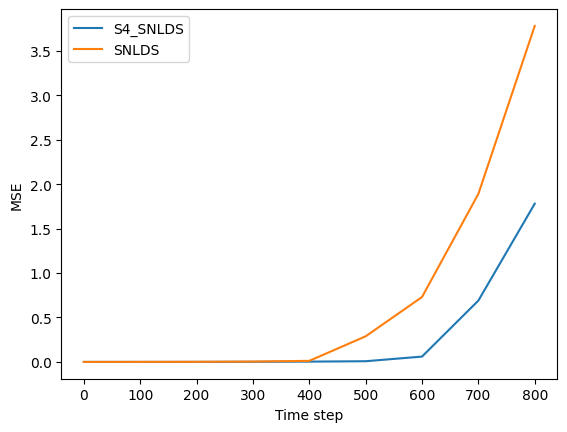}
    \caption{MSE against time step for generation task in 1-D nonlinear dynamics Lorenz attractor.}
    \label{fig:MSE-snlds}
\end{figure}

To understand the impact on accuracy over time, Fig. \ref{fig:MSE-snlds} plots the Mean Squared Error (MSE) against the generation timestep. Observing the trajectories in Fig. \ref{fig:show_lorenz_attractor_500}, we can gauge the S4 SNLDS model's prowess in emulating long-range dynamics inherent to nonlinear systems. Through Fig. \ref{fig:MSE-snlds}, we assess the span within which S4 SNLDS can predict accurately. Impressively, in our tests, it remains accurate up to 600 steps, surpassing the SNLDS, which caps at 400 steps.

\subsection{2-D bouncing ball}

The input data for 2-D bouncing ball experiment is a set of 100000 sample trajectories with 200 time steps. The initial position for the ball are randomly place between board with size $256 \times 256$. The velocity of the ball for each trajectories are randomly distributed by a uniform distribution $\mathcal{U}([-5,5])$. The training is performed with batch size 64. The test set is on a fixed 100 samples. The continues hidden state $\mathbf{z}$ is 10-D, and the discrete hidden state is 8-D. The observations pass through an encoder that contain two fully connected NN with activation function ReLU that contain 1-D convolution with 2 kernels of size 3. Then it goes to RNN inference network that both of the S4 RNN are 6 layers. The emission network is a 4-layer MLP with 3 layer of ReLU activation and a linear output. The Discrete hidden state transition network takes the previous discrete state and the processed observations as input. The continues hidden state  transition network is a single layer MLP with ReLU activation function. And the learning rate is constant $10^{-4}$. 

\begin{figure}
    \centering
    \includegraphics[width=1\linewidth]{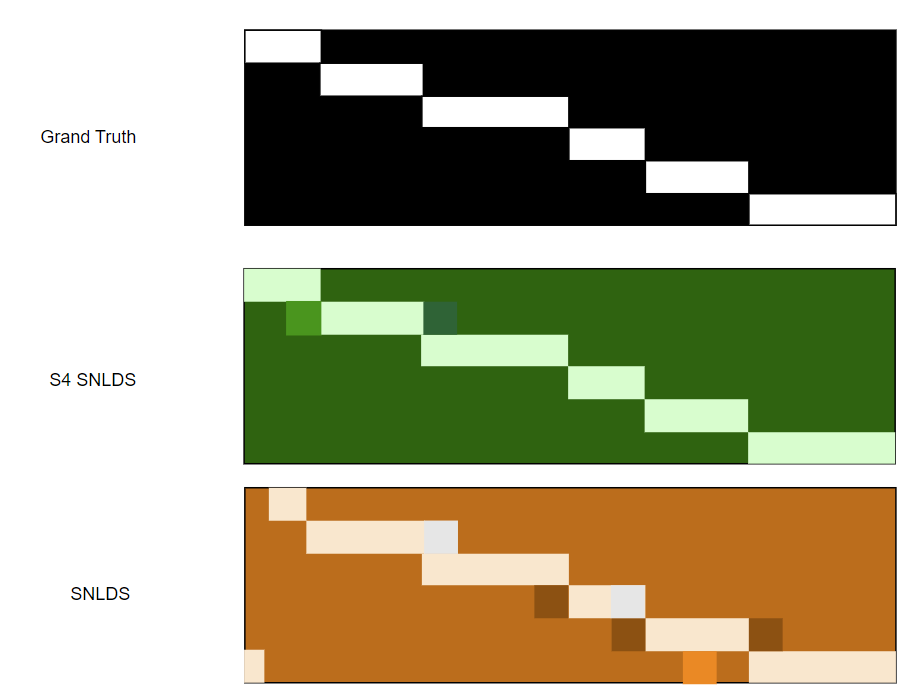}
    \caption{Segmentation on bouncing ball for 100 time steps. From top to bottom: ground truth of latent discrete, the posterior marginals of S4 SNLDS and SNLDS. Lighter colour represents higher probability states.}
    \label{Segmentation on bouncing ball}
\end{figure}

\begin{figure}
    \centering
    \includegraphics[width=1\linewidth]{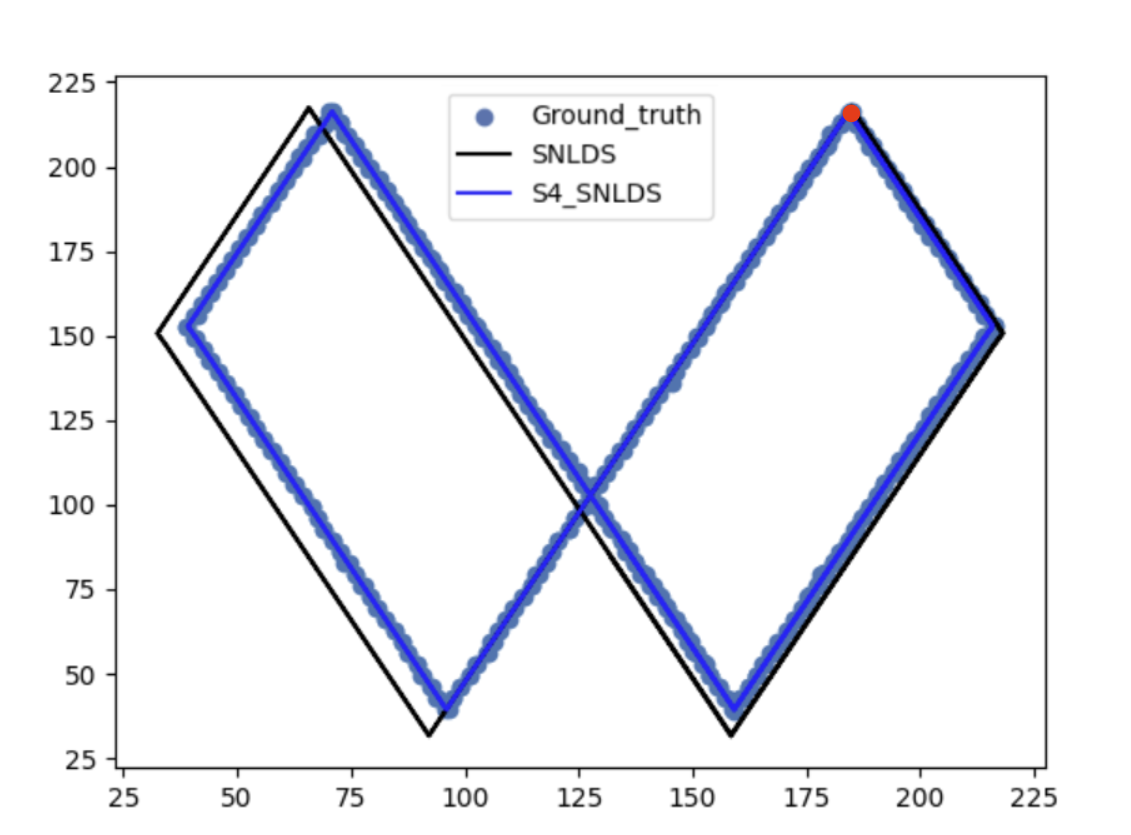}
    \caption{Bouncing sequence. Red dot means the starting point.}
    \label{fig:bouncing_sequence_compare_long-label}
\end{figure}

We applied both the S4 SNLDS and SNLDS models to this dataset, which presents a more complex challenge due to its intricate dynamics. Fig. \ref{Segmentation on bouncing ball} illustrates the segmentation results for a selected sample from the test set. Observably, our S4 SNLDS aligns more accurately with the ground truth compared to SNLDS. For a more direct performance comparison, we depicted the trajectory of the bouncing ball from a sample in Fig. \ref{fig:bouncing_sequence_compare_long-label}. As evident, post the second switch, the SNLDS lags in promptly adapting when the state alters, though it eventually redirects towards the correct trajectory.

\section{Conclusion}
\label{Conclusion}

In this paper, we introduced the S4 split method, an advanced approach adept at deciphering the switching behaviour in irregular time series with extended dependencies, notably outperforming the foundational S4 model in forecasting tasks. Nevertheless, the S4 split is confined to offline settings, meaning it doesn't recognise real-time changes during its prediction phase. To bridge this gap, we unveiled the S4 SNLDS model. This hybrid approach melds the strengths of the S4 and SNLDS models, enabling the segmentation of high-dimensional sequences into discrete models. Crucially, it informs the continuous state about potential mode changes. With its ability to recall the comprehensive input history and its corresponding state – all the while updating as time progresses – the S4 SNLDS stands out.

Our experimentation with the 1-D Lorenz example hints at the potential of S4 SNLDS in diverse scientific contexts, including multi-neuronal spike train analyses. Moreover, the inherent flexibility of the S4 model lends itself to broader applications, such as predicting medical datasets, including ECG data.

\end{document}